\documentclass[a4paper,fleqn]{cas-dc}

\usepackage[authoryear,longnamesfirst]{natbib}

\usepackage{amsmath,amssymb,amsfonts,bm}
\usepackage{graphicx}
\usepackage{booktabs,multirow,makecell,array}
\usepackage{colortbl}
\usepackage{subfig}
\usepackage{algorithm}
\usepackage{algorithmic}
\usepackage{pifont}
\usepackage{xcolor}

\newcolumntype{I}{!{\vrule width 2pt}}
\newlength\savedwidth

\newlength\savewidth

\definecolor{Gray}{gray}{0.93}
\definecolor{lightgray}{gray}{0.92}

\newcommand{\cmark}{\ding{51}}
\newcommand{\xmark}{\ding{55}}
\newcommand{\method}[0]{GradMAP}

\newif\ifshowdiff
\showdifffalse                 
\definecolor{revcolor}{RGB}{0,90,200}
\ifshowdiff
  \newcommand{\rev}[1]{{\color{revcolor}#1}}
  \newcommand{\revm}[1]{{\color{revcolor}#1}}
\else
  \newcommand{\rev}[1]{#1}
  \newcommand{\revm}[1]{#1}
\fi
\ifshowdiff
  \newcommand{\revcoloron}{\color{revcolor}}
  \newcommand{\revcoloroff}{\color{black}}
\else
  \newcommand{\revcoloron}{}
  \newcommand{\revcoloroff}{}
\fi

\begin{document}
\let\WriteBookmarks\relax
\def\floatpagepagefraction{1}
\def\textpagefraction{.001}

\shorttitle{GradMAP: Faster Layer Pruning with Gradient Metric and Projection Compensation}

\shortauthors{H. Liu et al.}

\title[mode = title]{GradMAP: Faster Layer Pruning with Gradient Metric and Projection Compensation}

\author[1,2]{Hao Liu}
\fnmark[1]
\ead{liuhao2025@ia.ac.cn}
\credit{Conceptualization, Methodology, Software, Writing - original draft}

\author[1,2]{Guangyan Li}
\fnmark[1]
\ead{liguangyan2022@ia.ac.cn}
\credit{Methodology, Software, Validation}

\author[3]{Wensheng Zhang}
\ead{zhangwenshengia@hotmail.com}
\credit{Supervision, Writing - review \& editing}

\author[1]{Yongqiang Tang}
\cormark[1]
\ead{yongqiang.tang@ia.ac.cn}
\credit{Supervision, Writing - review \& editing}

\affiliation[1]{organization={Institute of Automation, Chinese Academy of Sciences},
            city={Beijing},
            country={China}}

\affiliation[2]{organization={School of Artificial Intelligence, University of Chinese Academy of Sciences},
            city={Beijing},
            country={China}}

\affiliation[3]{organization={Guangzhou University},
            city={Guangzhou},
            country={China}}

\cortext[1]{Corresponding author}
\fntext[fn1]{H. Liu and G. Li contributed equally to this work.}

\begin{abstract}
Large Language Models (LLMs) exhibit strong reasoning abilities, but their high computational costs limit their practical deployment.
Recent studies reveal significant redundancy in LLMs layers, making layer pruning an active research topic.
Layer pruning research primarily focuses on two aspects: measuring layer importance and recovering performance after pruning. Unfortunately, the present works fail to simultaneously maintain pruning performance and efficiency.
In this study, we propose GradMAP, a faster layer pruning method with \textbf{Grad}ient \textbf{M}etric \textbf{A}nd \textbf{P}rojection compensation, which consists of two stages.
In the first stage, we introduce a novel metric based on gradient magnitudes, enabling a global assessment of layer importance. Note that, it requires only a single backward propagation step per pruning decision, substantially enhancing pruning efficiency.
In the second stage, we first analyze the layers with the largest mean shift resulting from pruning, and then incorporate a simple yet effective projection compensation matrix to correct this drift in one step. In this way, the degradation of model performance caused by layer pruning is effectively alleviated. Extensive experiments show that GradMAP outperforms previous layer pruning methods in both pruning speed (achieving an average $4\times$ speedup) and performance.
\end{abstract}



\maketitle


\section{Introduction}\label{sec:intro}

Large Language Models (LLMs)~\cite{brown2020language,achiam2023gpt,llama2,vicuna} have demonstrated remarkable capabilities across various domains~\cite{hendrycks2020measuring,chiang2024chatbot}.
However, deploying these models remains a significant challenge due to their huge computational and memory demands. 
Numerous techniques have been proposed to compress transformer-based models, including pruning \cite{magitude-prune,Cofi,OBC,ProtoNAS,WeightPruning,KBS-Pruning-1,KBS-Pruning-2}, low-rank approximation \cite{first-SVD,LoRAP,KBS-Low-Rank-3,KBS-Low-Ranks-4}, quantization \cite{GPTQ,zeroquant,QLORA,ArchitectureQuantifying}, and knowledge distillation \cite{Distillation,KBS-Quant-5}.
Quantization reduces memory usage but depends on specialized hardware. Knowledge distillation compresses models but requires additional training. Low-rank approximation reduces size but demands further optimization. In contrast, pruning removes redundant parameters without retraining, offering an effective way.

\begin{figure}
    \centering
    \includegraphics[width=3.4in]{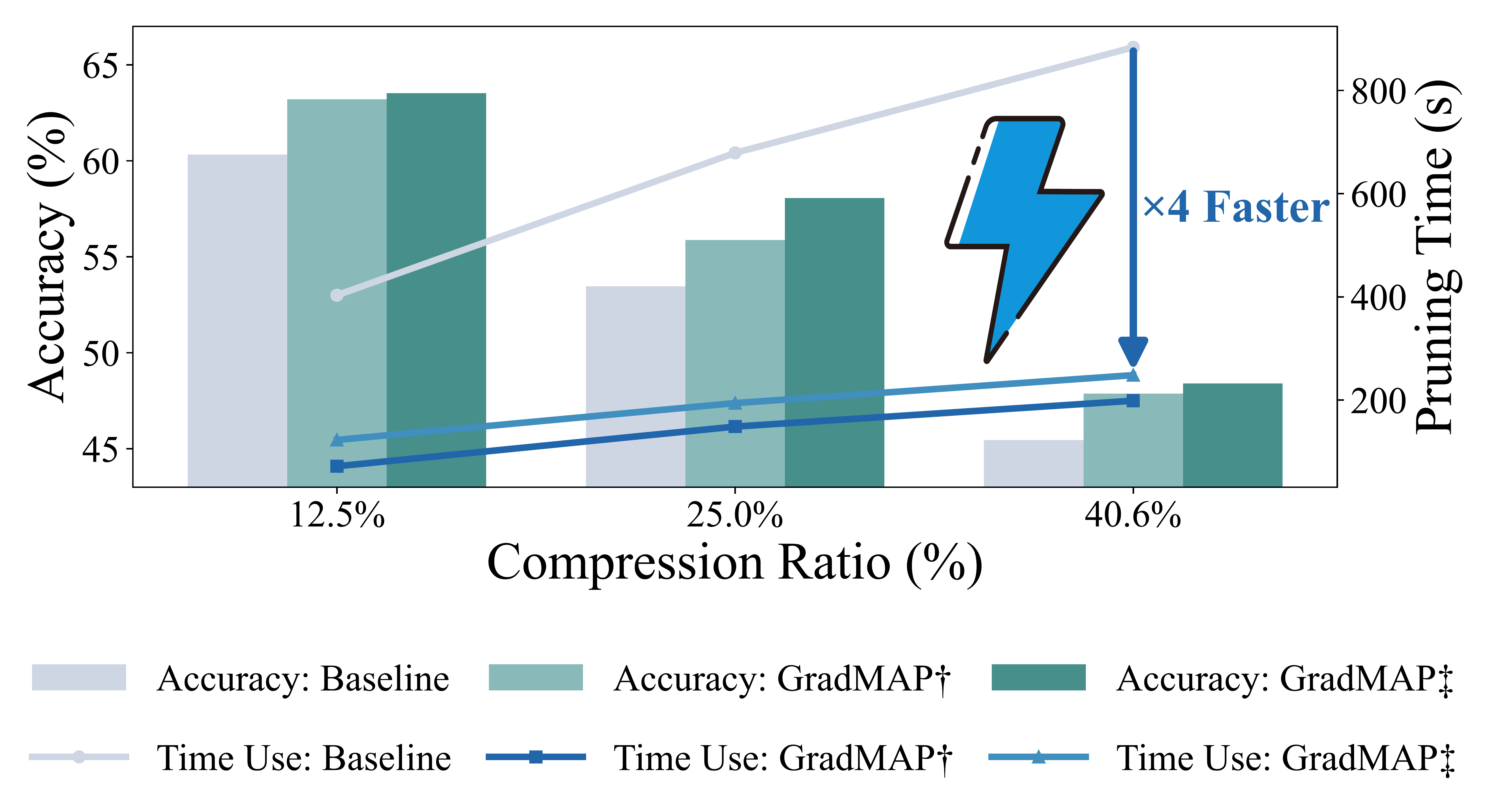}
    \caption[c]{Comparison of pruning time and accuracy across different methods on the Vicuna-7B.} 
    \label{fig:intro_comparison}
    \vspace{-0.5cm} 
\end{figure}


Previous work mainly focuses on pruning the dense matrices in LLMs, which leads to two main directions: unstructured pruning and structured pruning. 
Unstructured pruning methods, such as Wanda~\cite{wanda}, remove individual weights based on their importance. While effective, unstructured pruning modifies the internal weight matrices, making it more challenging to deploy pruned models in real-world systems. On the other hand, structured pruning methods, such as LLM-Pruner \cite{llm-pruner}, remove groups of neurons based on their connectivity, achieving significant model compression without compromising performance.
Building upon structured pruning, layer pruning further reduces model complexity by removing entire layers, rather than individual weights or neurons. 
This approach does not alter the internal weight structure, improving computational efficiency~\cite{SLEB} and easier to integrate into existing model pipelines~\cite{LLM-Streamline}. 
Recently, due to its direct impact on reducing parameter scale, layer pruning has become a popular focus in LLMs compression research.

\begin{figure*}[ht]
    \centering
    \includegraphics[width=\linewidth]{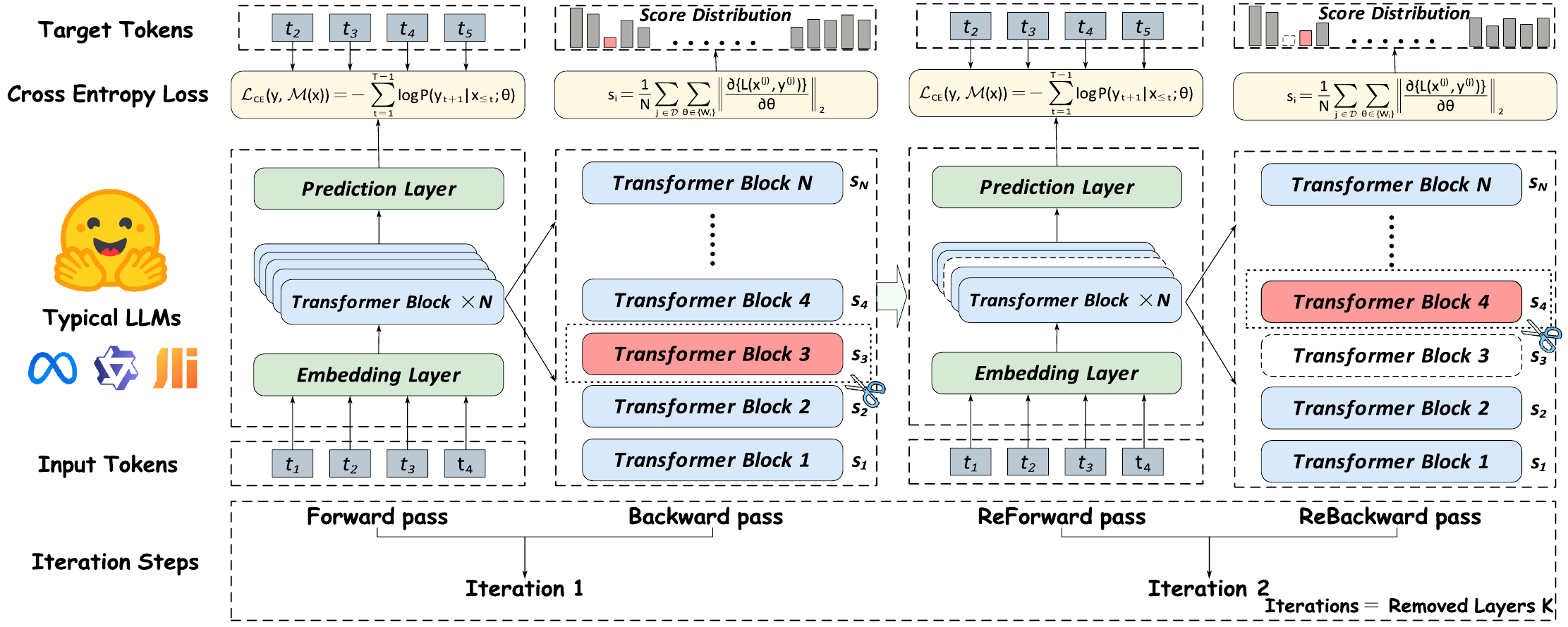}
    \vspace{-1mm}
    \caption{The Stage 1 of GradMAP introduces a novel gradient-based metric for estimating layer importance in LLMs, which quantifies each layer's contribution by analyzing gradient magnitudes. This metric enables the iterative identification and pruning of unimportant layers.}
    \label{fig:gradmap_stage1}
    \vspace{-1mm}
\end{figure*}

Existing layer pruning research can be viewed through two lenses: (1) how to quantify a layer’s importance; (2) how to compensate the performance lost once layers are pruned. As for the \emph{importance measurement} of layers, a straightforward idea is to compare a layer’s input and output representations with cosine similarity (e.g., ShortGPT~\cite{ShortGPT} and Laco~\cite{Laco}). Such metric only measures the similarity of hidden states and does not consider how a layer contributes to the task performance, thus leading to potential inaccuracies in identifying truly unimportant layers. To remedy this issue, SLEB~\cite{SLEB} and BlockPruner~\cite{BlockPruner} progressively mask each layer or sub-layer and then calculate the resulting loss change to identify the non-essential layers. However, their evaluation loop incurs substantial computational overhead, limiting their effectiveness and efficiency in practical pruning of LLMs.
It is worthy noting that, the methods mentioned above merely  remove unimportant layers in a rough manner, without any extra compensation mechanisms,  resulting in suboptimal performance. 
More recently, an increasing body of research has recognized the necessity of \emph{performance recovery} after layers pruning. For instance, LLM-Streamline \cite{LLM-Streamline} trains a lightweight layers to recover performance without significantly increasing the model size. UIDL~\cite{UIDL}, on the other hand, restores accuracy by combining pruning with conventional fine-tuning techniques such as QLoRA. While effective, these approaches typically rely on a large number of calibration samples and demand considerable computational resources and fine-tuning time, making them less efficient.

\begin{figure*}[t]
    \centering
    \includegraphics[width=\linewidth]{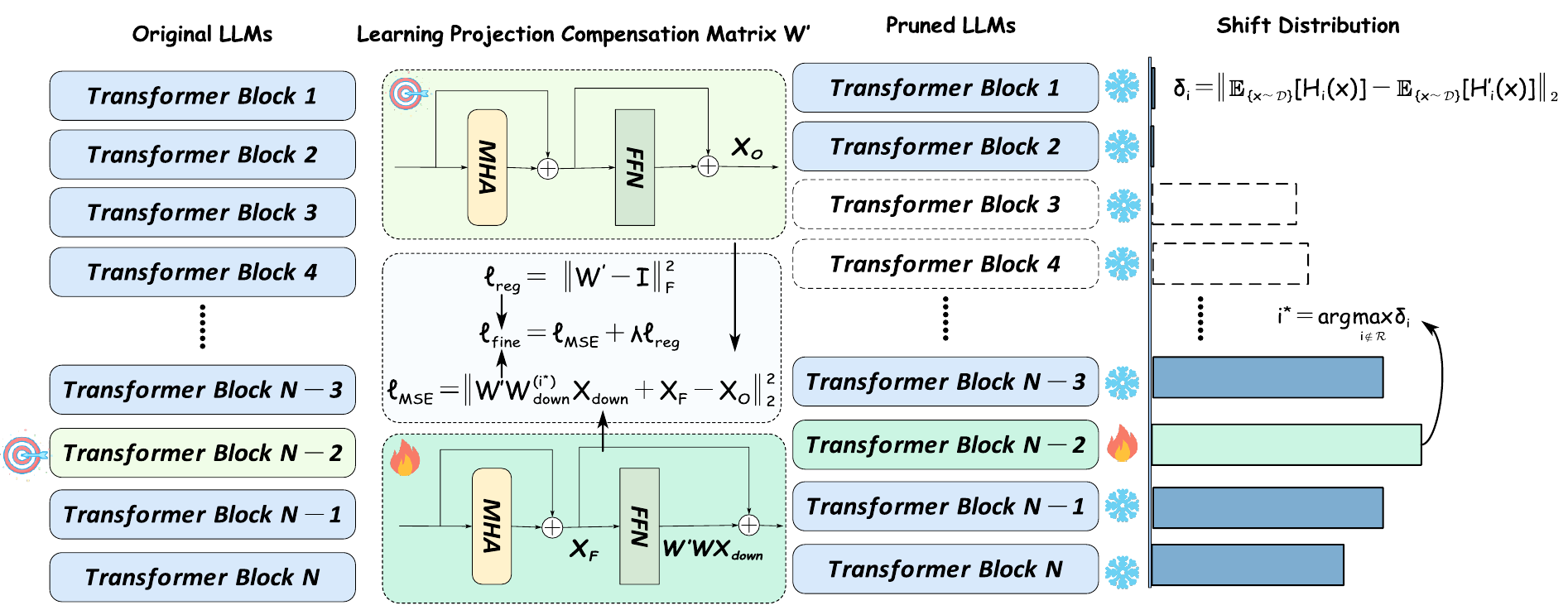}
    \vspace{-1mm}
    \caption{The Stage 2 of \method. To mitigate capacity loss, we introduce a learnable projection compensation matrix that optimizes reconstruction on a small calibration dataset.}
    \label{fig:gradmap_stage2}
    \vspace{-1mm}
\end{figure*}

To address the noted challenges, we propose GradMAP, a faster layer pruning method with \textbf{Grad}ient \textbf{M}etric \textbf{A}nd \textbf{P}rojection compensation.
Previous studies have demonstrated that gradient information inherently captures parameter sensitivity and the influence on task optimization~\cite{molchanov2016pruning,singh2020woodfisher}.
Building on this insight, we propose a layer importance metric based on global gradient magnitudes to directly quantify each layer’s contribution to model performance. Moreover, to enable rapid recovery of model performance after pruning, we analyze the shift in the after-pruning activations’ first-order moments and introduce a gradient-free projection compensation matrix. This matrix aligns the pruned model’s outputs with those of the original model, effectively compensating for the performance degradation with minimal computational overhead.
In summary, GradMAP operates in two stages: In the first stage, we introduce a novel importance metric based on global gradient magnitudes, allowing precise identification of unimportant layers. In the second stage, we analyze the shift in the first-order moment of activations caused by pruning and mitigate it by training a projection compensation matrix that aligns the after-pruning outputs with their original counterparts, thereby recovering performance with minimal overhead. 
As shown in Figure~\ref{fig:intro_comparison}, our method prunes faster and performs better than prior approaches.
Our contributions can be summarized as follows:

\begin{itemize}
\item We propose GradMAP, a faster and more effective layer pruning method. To better quantify the importance of each layer, we introduce a novel metric based on global gradient magnitudes. Additionally, we propose a projection compensation matrix to restore model performance after pruning.
\item The projection compensation matrix we propose can be seamlessly integrated into existing layer pruning frameworks, offering an effective means of recovering model performance while introducing minimal computational overhead. This integration can significantly enhance the pruning efficiency of current methods.
\item We evaluate the performance of the compressed model through zero-shot perplexity on the WikiText2, PTB and C4 datasets, as well as zero-shot task classification on common-sense reasoning datasets. Our method not only achieves an average pruning speed improvement of $4 \times$, but also outperforms existing layer pruning methods.
\end{itemize}

\section{Related Work}\label{Related Works}
In this section, we briefly review the most related works to ours, including traditional pruning methods, structured model compression, and recent advances in LLM pruning.
\subsection{Traditional Pruning}
Pruning has been proven to be an efficient approach for compressing pre-trained language models by removing redundant parameters~\cite{han2015deep_compression,molchanov2019importance,han2015learning,he2017channel,Partial,10521509}.
Traditionally, pruning methods focused on changing weight matrices, leading to two main categories: structured pruning and unstructured pruning. 
Structured pruning~\cite{llm-pruner,fang2024maskllm,zhang2023loraprune} removes entire neurons or attention heads, while unstructured pruning~\cite{wanda,frantar2023sparsegpt} sparsely eliminates individual weights. 
Despite their effectiveness, both approaches fundamentally alter the neural structure, posing significant challenges for downstream deployment due to hardware incompatibility and inference inefficiencies.

\subsection{Layer Pruning}
To address these limitations, layer pruning has emerged as a more flexible alternative. Recent  works on layer pruning have explored various strategies in terms of importance measurement and performance recovery.
For importance measurement, ShortGPT~\cite{ShortGPT} and Laco~\cite{Laco} use cosine similarity between input and output hidden states to detect redundancy, but such methods may overlook a layer’s functional contribution. Loss-based methods like SLEB~\cite{SLEB} and BlockPruner~\cite{BlockPruner} address this by masking layers and measuring loss increase, albeit with high computational cost due to repeated forward passes.
In summary, cosine similarity overlooks functional impact, while loss-based methods are computationally costly.
Moreover, most approaches remove or merge layers without compensating for performance loss.
Therefore, recent works like LLM-Streamline~\cite{LLM-Streamline} and UIDL~\cite{UIDL} introduce lightweight retraining or QLoRA-based tuning to restore accuracy, but these require substantial resources.
Hence, in this paper, we focus on both accurate layer importance measurement and efficient performance compensation techniques.

\section{Methodology}
\subsection{Problem Formulation}
Most Large Language Models (LLMs) are built on the Transformer architecture~\cite{vaswani2017attention}, where each layer consists of two key components: a self-attention layer and a feed-forward network (FFN) layer. 
Given the hidden representation $\mathbf{H}_{(i-1)}$ from layer $(i{-}1)$, the $i$-th layer’s transformation is:
\begin{equation}
    \mathbf{H}_{i} = \text{FFN}(\text{Self-Attn}(\mathbf{H}_{(i-1)})),
\end{equation}
where Self-Attn computes dependencies between tokens using the attention mechanism, and FFN applies non-linear transformations. The standard design stacks $L$ layers to refine representations.

Prior studies~\cite{SLEB,BlockPruner,ShortGPT} have demonstrated that LLMs contain considerable redundancy across layers. These findings have led to a common layer pruning paradigm: assess each layer’s importance, remove the unimportant ones, and optionally fine-tune the model to recover accuracy. In this study, we formalize layer pruning as:

\textbf{Importance Scoring}: Compute layer-wise importance scores:
\begin{equation}
    s_i = \Phi(\theta_{i}; \mathcal{D}),
\end{equation}
where $\theta_{i}$ can represent either model parameters~\cite{SLEB}) or hidden states $\mathbf{H}_{i}$~\cite{ShortGPT}, and $\mathcal{D}$ denotes a calibration dataset.

 \textbf{Adaptive Selection and Fine-tuning}: Determine a ranking permutation $\pi = \text{argsort}(s_i)$ such that $s_{\pi(1)} \geq s_{\pi(2)} \geq \dots \geq s_{\pi(L)}$. The pruned model retains the top $(L-K)$ layers:  
\begin{equation}
    \mathcal{S} = \{\pi(1), \pi(2), \dots, \pi(L-K)\},
\end{equation}
where \(L\) denotes the total number of layers, and \(K\) represents the number of pruned layers.

Optionally, fine-tune the pruned model on a small calibration dataset $\mathcal{D}$. The updated model parameters are obtained by minimizing the loss:
\begin{equation}
    \theta^* = \arg\min_{\theta} \mathcal{L}(\mathcal{D}; \theta).
\end{equation}

\subsection{The Proposed GradMAP}
\label{GradMAP}
Building on the standard layer pruning framework, we propose GradMAP. 
In Stage 1, GradMAP quantifies each layer's contribution to model performance by analyzing the gradients propagated through the network using calibration data. Unlike conventional methods that rely solely on hidden state similarity, this approach provides a more precise identification of redundant layers, ensuring that pruning minimally impacts model performance.
Furthermore, it is important to highlight that although our approach utilizes gradients, we do not perform any training or parameter updates on the LLM itself. 
In Stage 2, we introduce a projection compensation matrix to address the performance degradation caused by pruning. Notably, this stage does not require any further gradient computations, making it highly efficient in terms of both time and computational resources.

\textbf{Stage 1: Measuring Layer Importance via Gradient Magnitude.} 
Given a pre-trained language model $\mathcal{M}$ parameterized by $\theta$ with $L$ layers, processing the input sequence $\mathbf{x} = [x_1,...,x_T]$ and generating the target sequence $\mathbf{y} = [y_1,...,y_T]$. The cross-entropy loss is defined as:
\begin{equation}
\mathcal{L}_{\text{CE}}(\mathbf{x},\mathbf{y}) = -\sum_{t=1}^{T-1} \log P(y_{t+1}|\mathbf{x}_{\leq t};\theta).
\label{eq:loss}
\end{equation}

To estimate layer-wise importance, we utilize a small calibration dataset $\mathcal{D} = \{(\mathbf{x}^{(j)}, \mathbf{y}^{(j)})\}_{j=1}^N$ containing $N$ examples sampled from {Wikipedia}.
{Given the cross-entropy loss $\mathcal{L}_\text{CE}$ defined in Eq.~\eqref{eq:loss}, we first perform a forward pass on each input-output pair $(\mathbf{x}^{(j)}, \mathbf{y}^{(j)}) \in \mathcal{D}$ to compute the loss value $\mathcal{L}_\text{CE}(\mathbf{x}^{(j)}, \mathbf{y}^{(j)})$. }
And we then apply backpropagation to obtain the gradients of this loss with respect to all parameters in each layer.

Specifically, for each layer indexed by $i = 1, \ldots, L$, we define its trainable parameter set $\mathcal{W}_i$ to include all learnable tensors within the layer. For each calibration sample, we compute the total gradient \rev{energy} of layer index $i$ by summing the \rev{squared} $L_2$ norms of the gradients of all tensors $\theta \in \mathcal{W}_i$:
\begin{equation} G_i^{(j)} = \sum_{\theta \in \mathcal{W}_i} \left\| \frac{\partial \mathcal{L}_\text{CE}(\mathbf{x}^{(j)}, \mathbf{y}^{(j)})}{\partial \theta} \right\|_2^{\revm{2}}.
\label{eq:gradient}
\end{equation}

The final importance score for layer $i$ is then defined as the average of $G_i^{(j)}$ over all samples:
\begin{equation}
\begin{aligned}
s_i &=  \frac{1}{N} \sum_{j \in \mathcal{D}} \sum_{\theta \in \mathcal{W}_i} \left\| \frac{\partial \mathcal{L}_\text{CE}(\mathbf{x}^{(j)}, \mathbf{y}^{(j)})}{\partial \theta} \right\|_2^{\revm{2}} \\
&\revm{=\; \frac{1}{N} \sum_{j \in \mathcal{D}} \big\| \nabla_{\mathcal{W}_i}\mathcal{L}_\text{CE}(\mathbf{x}^{(j)}, \mathbf{y}^{(j)}) \big\|_2^{2}}.
\end{aligned}
\label{eq:importance score}
\end{equation}

This gradient-based formulation also admits a principled interpretation from the perspective of Fisher information.
Recall that the Fisher information of \rev{a single} parameter $\rev{\theta}$ is defined as
\begin{equation}
\mathcal{I}(\rev{\theta})=
\mathbb{E}_{(x,y)\sim\mathcal{D}}
\left[
\left(
\frac{\partial}{\partial\rev{\theta}}
\log p_\theta(y|x)
\right)^2
\right].
\end{equation}
Since the cross-entropy loss corresponds to the negative log-likelihood,
$\mathcal{L}_{CE}(x,y)=-\log p_\theta(y|x)$,
we have
\begin{equation}
\mathcal{I}(\rev{\theta})=
\mathbb{E}_{(x,y)\sim\mathcal{D}}
\left[
\left(
\frac{\partial
\mathcal{L}_{CE}(x,y)}
{\partial\rev{\theta}}
\right)^2
\right].
\end{equation}
\rev{Therefore, each squared gradient norm in Eq.~(\ref{eq:importance score}) is exactly the diagonal empirical Fisher information $\mathcal{I}(\theta)$ of parameter $\theta$. Summing over the tensor set $\mathcal{W}_i$, our score is by construction the trace of the layer’s block-diagonal Fisher,}
\begin{equation}
\revcoloron
s_i = \mathrm{tr}(F_i) = \sum_{\theta \in \mathcal{W}_i} \mathcal{I}(\theta),
\revcoloroff
\label{eq:fisher-trace}
\end{equation}
\rev{which quantifies the sensitivity of the model to layer $i$.}

\rev{To see why this trace serves as a valid importance metric, we derive an upper bound on the loss degradation caused by layer removal. In a residual transformer, each layer computes $\mathbf{H}_i = \mathbf{H}_{i-1} + f_i(\mathbf{H}_{i-1};\theta_i)$. Removing layer $i$ zeros its residual contribution, and by a first-order Taylor expansion the resulting loss change is}
\begin{equation}
\revcoloron
\Delta\mathcal{L} \approx -\nabla_{\mathbf{H}_i}\mathcal{L}^\top f_i.
\revcoloroff
\end{equation}
\rev{Applying Cauchy--Schwarz gives $|\Delta\mathcal{L}| \leq \|\nabla_{\mathbf{H}_i}\mathcal{L}\|_2 \cdot \|f_i\|_2$. Meanwhile, since layer $i$’s parameters only affect the loss through $f_i$, the chain rule yields $\nabla_{\theta_i}\mathcal{L} = J_i^\top \nabla_{\mathbf{H}_i}\mathcal{L}$, where $J_i = \partial f_i / \partial \theta_i$. Because $|\mathcal{W}_i| \gg d$ in practice, $J_i$ has full row rank, implying}
\begin{equation}
\revcoloron
\|\nabla_{\mathbf{H}_i}\mathcal{L}\|_2 \leq \frac{\|\nabla_{\theta_i}\mathcal{L}\|_2}{\sigma_{\min}(J_i)}.
\revcoloroff
\end{equation}
\rev{Let $M_i = \max_{\mathbf{x} \in \mathcal{D}} \|f_i(\mathbf{H}_{i-1}(\mathbf{x}))\|_2$. Combining and taking expectation over $\mathcal{D}$, we obtain}
\begin{equation}
\revcoloron
\mathbb{E}_{\mathcal{D}}\big[(\Delta\mathcal{L})^2\big]
\;\leq\;
\frac{M_i^2}{\sigma_{\min}^2(J_i)}
\;\cdot\; s_i.
\revcoloroff
\label{eq:loss-bound}
\end{equation}
\rev{The right-hand side is the product of our importance score $s_i$ and a layer-specific geometric constant $M_i^2/\sigma_{\min}^2(J_i)$ that does not depend on the pruning decision. Only $s_i$ varies across layers, so ranking by $s_i$ directly ranks layers by their worst-case removal impact.}

\rev{It is also worth noting a distinction from prior second-order pruning methods. Approaches such as Optimal Brain Damage~\cite{lecun1989optimal} estimate parameter importance via the quadratic form $\frac{1}{2}\theta^\top F \theta$, which requires computing or approximating the full Fisher matrix and often resorts to a diagonal approximation for tractability. In contrast, our metric uses the Fisher trace $\mathrm{tr}(F_i) = \|\nabla_{\mathcal{W}_i}\mathcal{L}\|_2^2$, which is computed exactly from gradient norms, requiring no matrix inversion or approximation. The validity of this layer-level aggregation is justified by the bound in Eq.~\eqref{eq:loss-bound}: the inequality involves the full gradient norm $s_i$, so the trace naturally captures the total sensitivity of the layer.}

As illustrated in Figure~\ref{fig:gradmap_stage1}, layers with lower importance scores contribute less to the model’s learning and thus can be pruned with minimal impact on performance. Specifically, we iteratively remove layers with lower importance scores until the target number of layers $L-K$ is reached.

\textbf{Stage 2: Performance Recovery via  Projection Compensation Matrix.}
To mitigate the performance degradation caused by pruned layers, GradMAP introduces a compensation mechanism. 
Specifically, we propose an adaptive compensation strategy that approximates the functionality of pruned layers using a learned transformation matrix applied to remaining layers.
This method can seamlessly integrates with existing pruning techniques, and enhances model performance with minimal computational overhead.

\begin{algorithm}[tb]
\caption{The pseudo-code of  GradMAP}
\label{alg:gradmap}
{
\raggedright
\textbf{Input}: Calibration dataset $\mathcal{D} = \{(\mathbf{x}^{(j)}, \mathbf{y}^{(j)})\}_{j=1}^{N}$, Original LLM $\mathcal{M}$, Target layer count $L-K$ \\
\textbf{Output}: Pruned LLM $\mathcal{M}'$ \par
}
\begin{algorithmic}[1]
\STATE \textbf{Stage 1: Gradient-Based Layer Pruning}
\STATE Initialize the set of removed layers \(\mathcal{R} = \emptyset\)
\WHILE{$|\mathcal{R}| < K$}
    \FOR{each layer $i \notin \mathcal{R}$}
        \STATE Compute the loss by Eq.(\ref{eq:loss});
        \STATE Compute the gradient by Eq.(\ref{eq:gradient});
        \STATE Compute the importance score by Eq.(\ref{eq:importance score});
    \ENDFOR
    \STATE Find worst layer: \(\ell^* = \arg\min s_{i}\);
    \STATE Update removal set: \(\mathcal{R} \leftarrow \mathcal{R} \cup \{\ell^*\}\);
    \STATE Remove layer $\ell^*$: $\mathcal{M} \leftarrow \text{Prune}(\mathcal{M}, \ell^*)$;
\ENDWHILE
\STATE \textbf{Stage 2: Projection Compensation Matrix Learning}
\STATE Compute the mean drift by Eq.(\ref{eq:mean-shift});
\STATE Identify the layer with largest drift by Eq.(\ref{eq:find-max-shift});
\STATE Extract the down-projection matrix \(\mathbf{W}_{down}^{(i^*)}\);
\STATE Learn $\mathbf{W}'$ by Eq.(\ref{eq:learning-obj});
\STATE Construct final model: $\mathcal{M}' \leftarrow \mathbf{W}'\mathbf{W}_{down}^{(i^*)}$.
\end{algorithmic}
\end{algorithm}

\begin{table*}[ht]
\centering
\footnotesize
\setlength{\tabcolsep}{4pt}
\renewcommand{\arraystretch}{1.2}
\resizebox{\textwidth}{!}{\begin{tabular}{c|c|ccc|ccc|ccc}
\toprule
\multirow{2}{*}{\textbf{Ratio}} & \multirow{2}{*}{\textbf{Method}} & \multicolumn{3}{c|}{\textbf{Average Accuracy$\uparrow$}} & \multicolumn{3}{c|}{\textbf{PPL$\downarrow$}} & \multicolumn{3}{c}{\textbf{Pruning Time$\downarrow$}} \\
& & \textbf{LLaMA2-7B} & \textbf{LLaMA2-13B} & \textbf{Vicuna-7B} & \textbf{LLaMA2-7B} & \textbf{LLaMA2-13B} & \textbf{Vicuna-7B} & \textbf{LLaMA2-7B} & \textbf{LLaMA2-13B} & \textbf{Vicuna-7B} \\
\midrule
\multirow{1}{*}{\textbf{0\%}} 
& Dense      & 66.79 & 69.31 & 66.85 & 12.18 & 10.98 & 16.23 & -- & -- & -- \\
\midrule
\multirow{7}{*}{\textbf{12.50\%}}
& LLM-Pruner    & 58.96 & 63.55 & 60.64 & 16.27 & 14.67 & 20.47 & -- & -- & -- \\
& Subcloning      & 59.92 & 61.25 & 61.48 & 24.26 & 19.91 & 27.55 & -- & -- & -- \\
& SLEB       & 60.87 & 60.60 & 62.79 & 16.68 & 14.63 & 21.68 & 503.98 & 1638.51 & 506.11 \\
& ShortGPT   & 62.32 & 64.14 & 63.21 & 16.85 & 14.78 & 21.63 & 519.33 & 1682.42 & 523.34 \\
& MKA        & 57.80 & 59.82 & 58.31 & 213.40 & 253.40 & 227.19 & 179.55 & 381.04 & 181.01 \\
& \cellcolor{lightgray}\textbf{\method\textsuperscript{\dag}}   & \cellcolor{lightgray}\textbf{62.54} & \cellcolor{lightgray}64.34 & \cellcolor{lightgray}63.21 & \cellcolor{lightgray}15.86 & \cellcolor{lightgray}14.78 & \cellcolor{lightgray}20.16 & \cellcolor{lightgray}\textbf{71.40} & \cellcolor{lightgray}\textbf{185.06} & \cellcolor{lightgray}\textbf{71.77} \\
& \cellcolor{lightgray}\textbf{\method\textsuperscript{\ddag}} & \cellcolor{lightgray}\textbf{62.54} & \cellcolor{lightgray}\textbf{64.39} & \cellcolor{lightgray}\textbf{63.52} & \cellcolor{lightgray}\textbf{15.49} & \cellcolor{lightgray}\textbf{14.55} & \cellcolor{lightgray}\textbf{20.12} & \cellcolor{lightgray}122.20 & \cellcolor{lightgray}261.65 & \cellcolor{lightgray}122.79 \\
\midrule
\multirow{7}{*}{\textbf{25.00\%}}
& LLM-Pruner       & 52.85 & 56.22 & 52.29 & 26.82 & 22.85 & 29.82 & -- & -- & -- \\
& Subcloning      & 53.85 & 52.44 & 55.09 & 59.06 & 55.24 & 63.62 & -- & -- & -- \\
& SLEB       & 55.96 & 57.79 & 54.88 & 21.76 & 18.83 & 31.66 & 801.58 & 2464.18 & 902.32 \\
& ShortGPT   & 54.19 & 58.97 & 52.98 & 33.31 & 23.85 & 48.37 & 831.16 & 2552.41 & 923.20 \\
& MKA        & 53.44 & 54.26 & 52.56 & 877.70 & 1252.39 & 1140.32 & 182.58 & 542.56 & 199.16 \\
& \cellcolor{lightgray}\textbf{\method\textsuperscript{\dag}}   & \cellcolor{lightgray}56.08 & \cellcolor{lightgray}59.33 & \cellcolor{lightgray}55.88 & \cellcolor{lightgray}21.50 & \cellcolor{lightgray}19.39 & \cellcolor{lightgray}28.49 & \cellcolor{lightgray}\textbf{123.42} & \cellcolor{lightgray}\textbf{299.30} & \cellcolor{lightgray}\textbf{148.63} \\
& \cellcolor{lightgray}\textbf{\method\textsuperscript{\ddag}} & \cellcolor{lightgray}\textbf{58.59} & \cellcolor{lightgray}\textbf{59.62} & \cellcolor{lightgray}\textbf{58.06} & \cellcolor{lightgray}\textbf{20.56} & \cellcolor{lightgray}\textbf{18.53} & \cellcolor{lightgray}\textbf{27.56} & \cellcolor{lightgray}169.42 & \cellcolor{lightgray}368.30 & \cellcolor{lightgray}194.13 \\
\midrule
\multirow{7}{*}{\textbf{40.63\%}}
& LLM-Pruner       & 41.08 & 42.09 & 41.06 & 103.21 & 90.37 & 120.86 & -- & -- & -- \\
& Subcloning      & 44.82 & 49.46 & 45.87 & 453.10 & 92.30 & 213.62 & -- & -- & -- \\
    & SLEB       & 42.40 & 46.77 & 46.74 & 103.61 & 44.65 & 85.23 & 1201.52 & 3411.58 & 1204.44 \\
& ShortGPT   & 44.87 & 49.34 & 42.46 & 384.95 & 60.67 & 649.73 & 1244.63 & 3530.80 & 1247.18 \\
& MKA        & {46.14} & 47.45 & 47.19 & 2682.50 & 3457.97 & 3325.50 & 197.08 & 564.52 & 211.60 \\
& \cellcolor{lightgray}\textbf{\method\textsuperscript{\dag}}   & \cellcolor{lightgray}46.26 & \cellcolor{lightgray}50.11 & \cellcolor{lightgray}47.88 & \cellcolor{lightgray}72.61 & \cellcolor{lightgray}42.77 & \cellcolor{lightgray}82.28 & \cellcolor{lightgray}\textbf{190.60} & \cellcolor{lightgray}\textbf{447.91} & \cellcolor{lightgray}\textbf{198.80} \\
& \cellcolor{lightgray}\textbf{\method\textsuperscript{\ddag}} & \cellcolor{lightgray}\textbf{46.72} & \cellcolor{lightgray}\textbf{50.39} & \cellcolor{lightgray}\textbf{47.93} & \cellcolor{lightgray}\textbf{56.99} & \cellcolor{lightgray}\textbf{39.17} & \cellcolor{lightgray}\textbf{80.06} & \cellcolor{lightgray}239.84 & \cellcolor{lightgray}522.06 & \cellcolor{lightgray}248.34 \\
\bottomrule
\end{tabular}}
\caption{Zero-shot evaluations of different pruning methods with 12.5\%, 25\%, and 40.63\% pruning ratios across various LLMs. `PPL' means perplexity on Wikitext2. 
\textbf{`bold'} indicates the best performance. \textsuperscript{\dag} denotes models pruned using only \textbf{Stage 1} (layer selection), while 
\textsuperscript{\ddag} indicates models further refined by \textbf{Stage 2} (projection compensation). Pruning time in seconds.} 
\label{tab:zeroshot}
\end{table*}

As illustrated in Figure~\ref{fig:gradmap_stage2}, we first quantify the drift in the first-order moment between the outputs of the retained layers before and after pruning. We then identify the layer exhibiting the largest drift and introduce a learnable projection compensation matrix \(W'\) to compensate the mismatch.
Formally, consider the original LLM \( \mathcal{M} \) with \(L\) layers indexed by $i = 1, \ldots, L$. After pruning a subset of layers with indices \( \mathcal{R} = \{r_1, r_2, \dots, r_K\}\), let \( \mathbf{H}_i(\mathbf{x}) \) and \( \mathbf{H}'_{i}(\mathbf{x}) \) denote the outputs of layer \( i \) in the original and pruned models, respectively. For each retained layer ($i \notin \mathcal{R}$), we quantify its mean drift before and after pruning on the calibration dataset $\mathcal{D}$ as:
\begin{equation}
\delta_i \;=\;
\Bigl\| \,
\mathbb{E}_{\mathbf{x} \sim \mathcal{D}}\bigl[ \mathbf{H}_i(\mathbf{x}) \bigr]
\;-\;
\mathbb{E}_{\mathbf{x} \sim \mathcal{D}}\bigl[ \mathbf{H}'_{i}(\mathbf{x}) \bigr]
\,\Bigr\|_2 .
\label{eq:mean-shift}
\end{equation}

Although it is possible to apply compensation to the Top-$Z$ layers with the largest drift, we find that increasing $Z$ does not consistently improve performance and instead unnecessarily increases computational overhead (see The Ablation Studies Section). Therefore, we adopt a simple yet effective approach by selecting only the largest drift layer for compensation, i.e., $Z=1$:
\begin{equation}
i^* = \arg\max_{i \notin \mathcal{R}} \delta_i.
\label{eq:find-max-shift}
\end{equation}

To compensate for this maximal drift, we first extract the down-projection matrix \(\mathbf{W}_{down}^{(i^*)} \in \mathbb{R}^{d \times k}\) from layer \( i^* \) that exhibits the largest drift.
Then, we introduce a projection compensation matrix \(\mathbf{W}' \in \mathbb{R}^{d \times d}\), trained specifically to align the after-pruning output distribution of layer \( i^* \) with its before-pruning counterpart. 
The combined optimization objective is defined as:
\begin{equation}
\mathbf{W}' = \arg\min_{\mathbf{W}'} 
\mathbb{E}_{(\mathbf{x},\mathbf{y})\sim\mathcal{D}}
\bigl[\mathcal{L}_\mathrm{MSE} + \mathcal{L}_\mathrm{reg}\bigr],
\label{eq:learning-obj}
\end{equation}

where the mean squared error (MSE) term is
\begin{equation}
\resizebox{0.4\textwidth}{!}{$
\mathcal{L}_\mathrm{MSE} =
\left\|
\mathbf{W}' \mathbf{W}_{down}^{(i^*)} \mathbf{X}_{down} 
+ \mathbf{X}_F - \mathbf{X}_O
\right\|_2^2
$}
\label{eq:mse-loss}
\end{equation}

and the regularization term is
\begin{equation}
\mathcal{L}_\mathrm{reg} =
\lambda \left\| \mathbf{W}' - \mathbf{I} \right\|_F^2.
\label{eq:reg-loss}
\end{equation}

Here, \(\mathbf{X}_O\) denotes the output activations before pruning, 
\(\mathbf{X}_F\) represents the input activations to the FFN sub-layer after pruning, 
and \(\mathbf{X}_{down}\) is the input to the down-projection matrix in the FFN sub-layer after pruning. 
The regularization term constrains \(\mathbf{W}'\) toward the identity matrix, minimizing distortion of the original representations.

\begin{table}[t]
\centering
\resizebox{0.48\textwidth}{!}{
\begin{tabular}{c|c|c|cc}
\toprule
\textbf{Type} & \textbf{Method} & \textbf{Ratio} & \textbf{Throughput (Tokens/s)} & \textbf{Latency (ms)} \\
\midrule
-- 
& Dense
& 0\%
& 299 (1.00$\times$)
& 1718.4 (1.00$\times$) \\
\midrule

\multirow{2}{*}{\textbf{Unstructured}}
& Wanda
& 50\%
& 293 (0.98$\times$)
& 1555.5 (1.10$\times$) \\
& SparseGPT
& 50\%
& 293 (0.98$\times$)
& 1555.5 (1.10$\times$) \\
\midrule

\multirow{4}{*}{\textbf{Structured}}
& LLM-Pruner
& 20\%
& 314 (1.05$\times$)
& 1534.3 (1.12$\times$) \\

& SliceGPT
& 20\%
& 314 (1.05$\times$)
& 1658.7 (1.04$\times$) \\

& SliceGPT
& 25\%
& 331 (1.11$\times$)
& 1440.7 (1.19$\times$) \\

& SliceGPT
& 30\%
& 343 (1.15$\times$)
& 1364.2 (1.26$\times$) \\
\midrule

\cellcolor{lightgray}\multirow{1}{*}{\textbf{Layer}}
& \cellcolor{lightgray}\textbf{\method\textsuperscript{\ddag}}
& \cellcolor{lightgray}\textbf{20\%}
& \cellcolor{lightgray}\textbf{381 (1.27$\times$)}
& \cellcolor{lightgray}\textbf{1364.1 (1.26$\times$)} \\
\bottomrule
\end{tabular}}
\caption{Throughput and latency comparison under different pruning types on LLaMA2-7B.}
\label{tab:throughput_latency}
\end{table}

For efficient deployment and inference, the calibrated down-projection weight matrix $\hat{\mathbf{W}}_{down}^{(i^*)}$ is then obtained by re-parameterizing the original weights $\mathbf{W}_{down}^{(i^*)}$ as:
\begin{equation}
\hat{\mathbf{W}}_{down}^{(i^*)} = \mathbf{W}' \mathbf{W}_{down}^{(i^*)}.
\end{equation}

\begin{figure}
\vspace{-10pt}
\centering
\subfloat[PPL on WikiText2]{
    \includegraphics[width=0.23\textwidth]{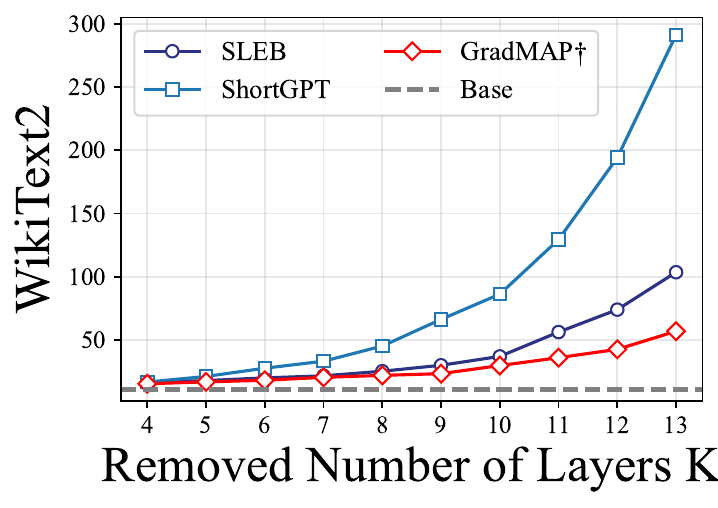}
}
\subfloat[PPL on PTB]{
    \includegraphics[width=0.23\textwidth]{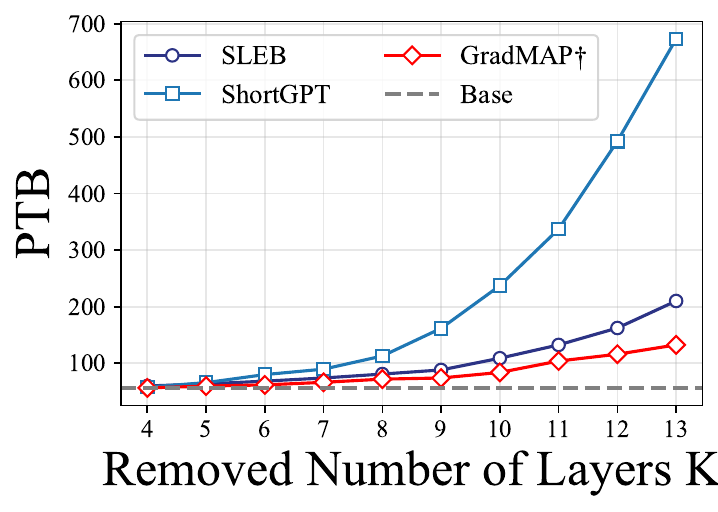}
}
\caption{Perplexity of LLaMA2-7B under varying compression ratios.}
\label{fig:diffent-ratio}
\end{figure}

\rev{The motivation behind this compensation design is as follows. Layer pruning removes entire computational blocks from the network. Even when the pruned layers are individually unimportant, their cumulative removal shifts the activation distribution of downstream layers. If left uncorrected, this distributional mismatch propagates and degrades the final output. We measure the first-order moment shift $\delta_i$ as defined in Eq.~\ref{eq:mean-shift} at every retained layer and find that the drift is highly concentrated: typically one layer absorbs the majority of the distributional mismatch, as visualized in Figure~\ref{fig:shift_opt} and Figure~\ref{fig:shift_llama2_vicuna}. Compensating additional layers yields diminishing returns and risks overfitting the limited calibration data, as confirmed by the ablation results in Table~\ref{tab:ablation_stage1_stage2}. Targeting only the most-drifted layer therefore offers the best tradeoff between recovery quality and computational cost.}

\rev{We apply the correction through the down-projection matrix $\mathbf{W}_{down}$ because it is the last linear operation in the FFN and maps directly to the layer's output space. The other weight matrices $\mathbf{W}_{up}$ and $\mathbf{W}_{gate}$ are followed by non-linear activations, so a linear correction applied to them would be distorted before reaching the output. Moreover, a learned projection matrix $\mathbf{W}'$ applied to $\mathbf{W}_{down}$ can be absorbed via re-parameterization as $\hat{\mathbf{W}}_{down} = \mathbf{W}'\mathbf{W}_{down}$, introducing zero inference overhead. Table~\ref{tab:ablation_weight_matrix} validates this design choice.}

\begin{figure}
    \centering
    \includegraphics[width=\linewidth]{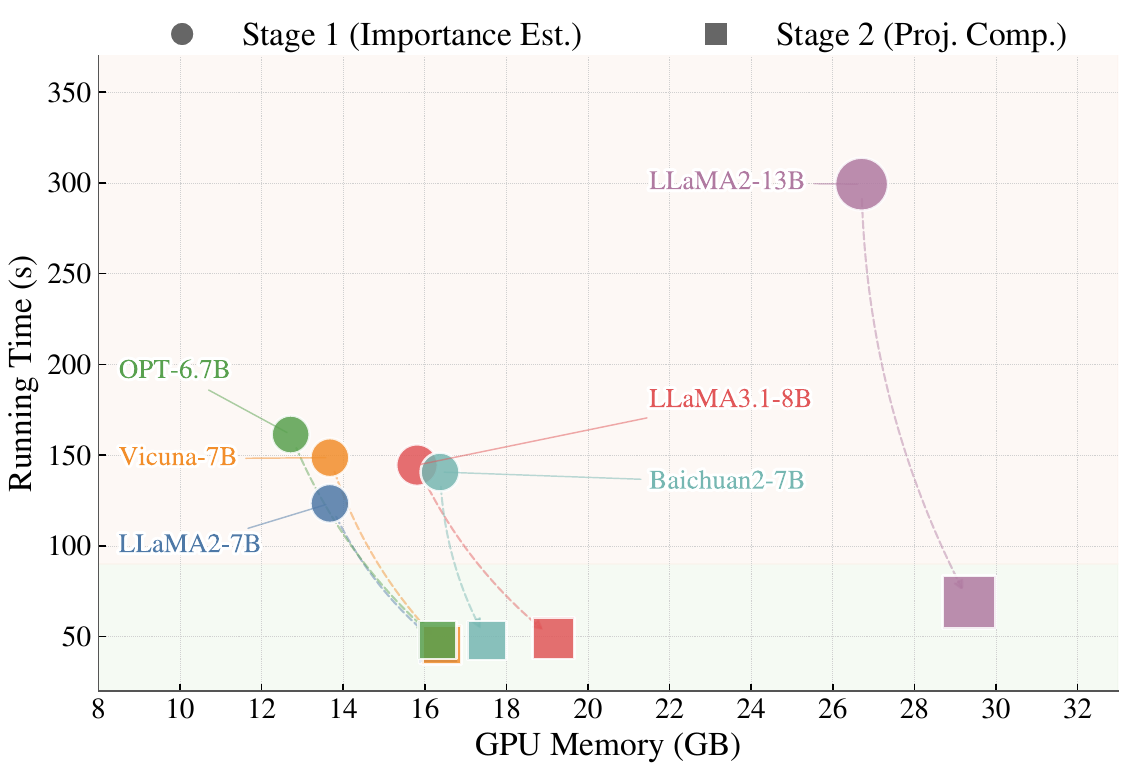}
    \vspace{-1mm}
    \caption{GPU memory consumption and running time of GradMAP during Stage 1 (importance estimation) and Stage 2 (projection compensation) under a 25\% compression ratio.}
    \label{fig:memory_time}
    \vspace{-1mm}
\end{figure}

\rev{The objective in Eq.~\eqref{eq:learning-obj} is convex and well-conditioned, making it straightforward to optimize. We solve it via gradient descent with the Adam optimizer, which converges rapidly since the problem involves only a single $d \times d$ matrix with a quadratic loss. The identity-regularization term biases $\mathbf{W}'$ toward $\mathbf{I}$, preventing overfitting on the limited calibration data and preserving the pre-trained representations. As shown in Figure~\ref{fig:memory_time}, this optimization typically converges within a few thousand steps and completes in under one minute on a single GPU.}

\begin{table*}[ht]
    \centering
    \renewcommand\arraystretch{1.2}
    \resizebox{\linewidth}{!}{
    \begin{tabular}{c|c|c|cccccccccccc|c}
        \toprule
        \multirow{2}{*}{Model} & \multirow{2}{*}{Method} & \multirow{2}{*}{Ratio} & \multicolumn{12}{c|}{Benchmarks} & \multirow{2}{*}{Average} \\
        & & & C3 & CMNLI & CHID & BoolQ & WSC & HeSW & PIQA & CoQA & Race-M & Race-H & MMLU & CMMLU & \\
        \midrule
        \multirow{5}{*}{LLaMA3.1-8B}
        & Dense  & 0.00\% & 54.08 & 32.98 & 36.99 & 69.79 & 71.10 & 74.66 & 80.96 & 71.42 & 70.61 & 63.15 & 66.74 & 50.87 & 59.07 \\
        & SLEB & 25.0\% & 37.15 & 33.10 & 11.79 & \underline{54.43} & \underline{62.50} & 48.96 & 70.46 & 39.72 & 21.31 & 21.38 & 25.29 & 25.29 & 37.62 \\
        & ShortGPT & 25.0\% & 43.23 & \underline{33.13} & 7.09 & \textbf{57.86} & \textbf{63.50} & 55.71 & 69.21 & 54.22 & 35.17 & \underline{35.17} & \textbf{56.69} & 35.12 & 45.50 \\
        & \cellcolor{lightgray}\textbf{\method\textsuperscript{\dag}}  & \cellcolor{lightgray}25.0\% & \cellcolor{lightgray}\underline{47.07} & \cellcolor{lightgray}32.72 & \cellcolor{lightgray}\underline{26.07} & \cellcolor{lightgray}53.15 & \cellcolor{lightgray}\textbf{63.50} & \cellcolor{lightgray}\underline{58.54} & \cellcolor{lightgray}\textbf{73.45} & \cellcolor{lightgray}\textbf{59.13} & \cellcolor{lightgray}\underline{40.60} & \cellcolor{lightgray}\textbf{35.48} & \cellcolor{lightgray}54.98 & \cellcolor{lightgray}\underline{39.60} & \cellcolor{lightgray}\underline{48.68} \\
        & \cellcolor{lightgray}\textbf{\method\textsuperscript{\ddag}} & \cellcolor{lightgray}25.0\% & \cellcolor{lightgray}\textbf{47.67} & \cellcolor{lightgray}\textbf{33.43} & \cellcolor{lightgray}\textbf{26.97} & \cellcolor{lightgray}53.79 & \cellcolor{lightgray}\underline{62.50} & \cellcolor{lightgray}\textbf{60.05} & \cellcolor{lightgray}\underline{72.91} & \cellcolor{lightgray}\underline{58.72} & \cellcolor{lightgray}\textbf{40.74} & \cellcolor{lightgray}33.85 & \cellcolor{lightgray}\underline{55.63} & \cellcolor{lightgray}\textbf{40.77} & \cellcolor{lightgray}\textbf{48.92} \\
        \midrule
        \multirow{5}{*}{Baichuan2-7B}
        & Dense  & 0.00\% & 62.41 & 33.28 & 9.59 & 62.94 & 66.65 & 64.33 & 74.59 & 66.00 & 50.77 & 52.20 & 54.68 & 56.91 & 53.20 \\
        & SLEB & 25.0\% & 39.18 & 31.70 & \underline{11.59} & 38.29 & 60.58 & 30.59 & 57.73 & 22.28 & 21.38 & 25.27 & 25.42 & 22.36 & 32.20 \\
        & ShortGPT & 25.0\% & \underline{47.45} & \underline{33.53} & 5.49 & \textbf{54.71} & \textbf{65.38} & 46.56 & 62.68 & \textbf{23.26} & \textbf{24.50} & 27.00 & 28.59 & 40.62 & 38.31 \\
        & \cellcolor{lightgray}\textbf{\method\textsuperscript{\dag}}  & \cellcolor{lightgray}25.0\% & \cellcolor{lightgray}46.58 & \cellcolor{lightgray}33.37 & \cellcolor{lightgray}10.94 & \cellcolor{lightgray}\underline{43.88} & \cellcolor{lightgray}\underline{63.46} & \cellcolor{lightgray}\underline{50.53} & \cellcolor{lightgray}\textbf{66.49} & \cellcolor{lightgray}22.91 & \cellcolor{lightgray}23.84 & \cellcolor{lightgray}\textbf{28.92} & \cellcolor{lightgray}\textbf{30.77} & \cellcolor{lightgray}\underline{48.89} & \cellcolor{lightgray}\underline{39.21} \\
        & \cellcolor{lightgray}\textbf{\method\textsuperscript{\ddag}} & \cellcolor{lightgray}25.0\% & \cellcolor{lightgray}\textbf{49.70} & \cellcolor{lightgray}\textbf{33.64} & \cellcolor{lightgray}\textbf{12.89} & \cellcolor{lightgray}41.96 & \cellcolor{lightgray}62.50 & \cellcolor{lightgray}\textbf{51.16} & \cellcolor{lightgray}\underline{66.21} & \cellcolor{lightgray}\underline{23.12} & \cellcolor{lightgray}\underline{24.47} & \cellcolor{lightgray}\underline{27.54} & \cellcolor{lightgray}\underline{30.17} & \cellcolor{lightgray}\textbf{51.84} & \cellcolor{lightgray}\textbf{39.60} \\
        \midrule
        \multirow{5}{*}{OPT-6.7B}
        & Dense  & 0.00\% & 37.86 & 32.83 & 13.14 & 64.04 & 66.35 & 62.77 & 75.73 & 56.92 & 25.63 & 25.81 & 24.82 & 25.33 & 41.04 \\
        & SLEB & 25.0\% & 35.56 & 32.80 & 0.00 & 45.63 & 63.46 & \textbf{50.88} & \textbf{72.25} & \textbf{42.18} & 22.77 & \underline{23.79} & 24.40 & \textbf{25.25} & 36.58 \\
        & ShortGPT & 25.0\% & 25.97 & 32.89 & 0.05 & 45.14 & 56.73 & 26.33 & 53.10 & 17.61 & 22.28 & 22.16 & 23.63 & \underline{25.24} & 29.26 \\
        & \cellcolor{lightgray}\textbf{\method\textsuperscript{\dag}}  & \cellcolor{lightgray}25.0\% & \cellcolor{lightgray}\underline{36.11} & \cellcolor{lightgray}\underline{32.96} & \cellcolor{lightgray}\underline{2.90} & \cellcolor{lightgray}\underline{59.76} & \cellcolor{lightgray}\underline{64.42} & \cellcolor{lightgray}49.36 & \cellcolor{lightgray}68.01 & \cellcolor{lightgray}\underline{38.82} & \cellcolor{lightgray}\textbf{25.63} & \cellcolor{lightgray}\textbf{24.24} & \cellcolor{lightgray}\underline{24.67} & \cellcolor{lightgray}25.03 & \cellcolor{lightgray}\underline{37.66} \\
        & \cellcolor{lightgray}\textbf{\method\textsuperscript{\ddag}} & \cellcolor{lightgray}25.0\% & \cellcolor{lightgray}\textbf{36.16} & \cellcolor{lightgray}\textbf{32.97} & \cellcolor{lightgray}\textbf{7.09} & \cellcolor{lightgray}\textbf{60.61} & \cellcolor{lightgray}\textbf{65.38} & \cellcolor{lightgray}\underline{49.78} & \cellcolor{lightgray}\underline{68.39} & \cellcolor{lightgray}38.33 & \cellcolor{lightgray}\underline{25.49} & \cellcolor{lightgray}23.38 & \cellcolor{lightgray}\textbf{24.74} & \cellcolor{lightgray}24.87 & \cellcolor{lightgray}\textbf{38.10} \\
        \bottomrule
    \end{tabular}
    }
    \caption{Accuracy of different pruning methods across classification benchmarks on LLaMA3.1-8B, Baichuan2-7B, and OPT-6.7B.}
    \label{tab:new_model}
\end{table*}

\textbf{Discussion.} The optimization procedure of GradMAP is summarized in Algorithm~\ref{alg:gradmap}. Compared to prior methods, GradMAP achieves superior efficiency from both the pruning and compensation stages. Firstly, the proposed importance metric in Stage 1 significantly reduces computational complexity compared to existing pruning methods. 
GradMAP uses gradient magnitude as a direct and differentiable metric for importance, enabling each pruning decision to be made with just one forward-backward pass. Hence, our method scales linearly rather than quadratically, dramatically accelerating the pruning procedure. {Secondly}, Stage 2 introduces a simple yet effective projection compensation matrix whose training is inherently lightweight. 
Unlike previous methods that require compensating or fine-tuning every layer, our projection matrix targets only the layer with the largest activation drift after pruning. This focused approach substantially reduces computational requirements, further enhancing the overall efficiency. 

\section{Experimental Setup}
\label{Experiments}
\label{sec:main_memory}

We evaluate GradMAP on several widely used large language models, including LLaMA2-7B and LLaMA2-13B~\cite{llama2}, Vicuna-7B~\cite{vicuna}, LLaMA3.1-8B~\cite{llama3}, Baichuan2-7B~\cite{baichuan2}, and Qwen2.5-7B~\cite{qwen2.5}. 
All models are obtained from publicly available Hugging Face implementations. 
During the compression process, we randomly sample 128 sequences from Wikipedia as calibration data, where each sequence contains 128 tokens.

For perplexity evaluation, we adopt three standard language modeling benchmarks: PTB~\cite{PTB}, WikiText-2~\cite{WikiText2}, and C4~\cite{C4Dataset}. 
Following common zero-shot evaluation protocols, we further evaluate the models on several commonsense reasoning tasks, including BoolQ~\cite{boolq}, PIQA~\cite{piqa}, HellaSwag (HellaS.)~\cite{Hellaswag}, WinoGrande (WinoG.)~\cite{Winogrande}, ARC-easy~\cite{ARC}, ARC-challenge~\cite{ARC}, and OpenBookQA~\cite{OBQA}. 
For calibration, we use 128 samples with a sequence length of 128 tokens, and the optimization adopts Adam with a learning rate of $1\times10^{-3}$ and a regularization weight of $1\times10^{-3}$. 
After compression, perplexity is computed on 128-token segments from WikiText-2, PTB, and C4, and zero-shot performance is evaluated using the lm-evaluation-harness framework~\cite{lm-evaluation-harness}. 
All compression experiments are conducted on two NVIDIA A40 GPUs with 48GB memory.

\section{Main Results}

\begin{table*}[t]
  \centering
  \setlength\tabcolsep{6pt}
  \footnotesize
  \resizebox{\textwidth}{!}{\begin{tabular}{c|l|ccccccc|c|c}
    \toprule
    \textbf{Model} & \multicolumn{1}{c|}{\textbf{Method}}  & \textbf{BoolQ}$\uparrow$ & \textbf{PIQA}$\uparrow$ & \textbf{HellaS.}$\uparrow$ & \textbf{WinoG.}$\uparrow$ & \textbf{ARC-e}$\uparrow$ & \textbf{ARC-c}$\uparrow$ & \textbf{OBQA}$\uparrow$ & \textbf{Avg.}$\uparrow$ & \textbf{$\Delta$} \\
    \midrule
    \multirow{5}{*}{\textbf{LLaMA2-7B}} 
    & SLEB & 61.38 & 72.25 & 62.27 & 60.85 & 63.34 & 33.53 & 38.00 & 55.96 & \multirow{2}{*}{\textcolor{blue}{+0.79}} \\
    & \quad w/ Ours Stage2 & 67.55 & 71.98 & 62.30 & 60.69 & 63.05 & 33.87 & 37.80 & 56.75 &  \\
    \cmidrule{2-11}
    & ShortGPT & 53.64 & 68.17 & 62.50 & 65.98 & 55.77 & 34.64 & 38.60 & 54.19 & \multirow{2}{*}{\textcolor{blue}{+0.84}} \\
    & \quad w/ Ours Stage2 & 58.62 & 68.44 & 62.15 & 65.59 & 56.78 & 34.81 & 38.80 & 55.03 &  \\
    \cmidrule{2-11}
    & \multicolumn{1}{>{\cellcolor{lightgray}}c|}{\textbf{\method\textsuperscript{\ddag}}} & \cellcolor{lightgray}\textbf{70.46} & \cellcolor{lightgray}\textbf{70.95} & \cellcolor{lightgray}\textbf{63.91} & \cellcolor{lightgray}\textbf{65.35} & \cellcolor{lightgray}\textbf{64.02} & \cellcolor{lightgray}\textbf{37.46} & \cellcolor{lightgray}\textbf{38.00} & \cellcolor{lightgray}\textbf{58.59} & \cellcolor{lightgray}-- \\
    \midrule
    \multirow{5}{*}{\textbf{Vicuna-7B}} 
    & SLEB & 66.64 & 70.29 & 60.17 & 55.96 & 62.54 & 33.79 & 34.80 & 54.88 & \multirow{2}{*}{\textcolor{blue}{+0.09}} \\
    & \quad w/ Ours Stage2 & 67.37 & 70.35 & 59.53 & 56.27 & 62.25 & 33.62 & 35.40 & 54.97 &  \\
    \cmidrule{2-11}
    & ShortGPT & 64.86 & 66.70 & 57.43 & 63.14 & 53.49 & 34.22 & 31.00 & 52.98 & \multirow{2}{*}{\textcolor{blue}{+0.40}} \\
    & \quad w/ Ours Stage2 & 64.80 & 66.32 & 57.40 & 64.33 & 54.25 & 34.56 & 32.00 & 53.38 &  \\
    \cmidrule{2-11}
    & \multicolumn{1}{>{\cellcolor{lightgray}}c|}{\textbf{\method\textsuperscript{\ddag}}} & \cellcolor{lightgray}\textbf{72.05} & \cellcolor{lightgray}\textbf{70.89} & \cellcolor{lightgray}\textbf{61.36} & \cellcolor{lightgray}\textbf{61.64} & \cellcolor{lightgray}\textbf{63.05} & \cellcolor{lightgray}\textbf{38.65} & \cellcolor{lightgray}\textbf{38.80} & \cellcolor{lightgray}\textbf{58.06} & \cellcolor{lightgray}-- \\
    \bottomrule
  \end{tabular}}
    \caption{Performance of integrating our projection compensation matrix with existing layer pruning methods on LLaMA2-7B and Vicuna-7B under 25\% compression ratio. \textcolor{blue}{$\Delta$} indicates improvement after applying our Stage 2.}
  \vskip -0.1in
  \label{tab:acc_withstage2}
\end{table*}

\textbf{Baselines.} We compare our method with the following strong layer-wise compression methods:

\rev{\textbf{1) LLM-Pruner}~\cite{llm-pruner} performs structured pruning by removing coupled neurons and attention heads based on Taylor-expansion importance estimation. It prunes the width (channel dimension) of each layer rather than removing entire layers, achieving compression through reduced hidden dimensions.}

\rev{\textbf{2) Subcloning}~\cite{subcloning} removes layers based on relative magnitude, a metric that measures the ratio between each layer's output norm and input norm. Layers with the smallest relative magnitude are identified as redundant and removed as a contiguous block.}

\textbf{3) SLEB}~\cite{SLEB} prunes redundant blocks by leveraging output similarity between adjacent layers. During the process, each layer is sequentially masked out, and the loss metrics are calculated to evaluate its impact on model performance. 
This iterative layer-wise masking continues until the desired sparsity level is achieved.

\textbf{4) ShortGPT}~\cite{ShortGPT} employs a Block Influence (BI) metric to measure input-output similarity, identifying and removing less influential layers. The BI metric is computed for each layer of the LLM, systematically removing layers with lower scores until the model reaches the target pruning size.

\textbf{5) MKA}~\cite{MKA} merges similar layers via manifold learning and Normalized Pairwise Information Bottleneck alignment. Instead of direct removal, it consolidates structural knowledge, enabling hardware-friendly compression with minimal performance loss.

\begin{figure}
    \centering
    \includegraphics[width=\linewidth]{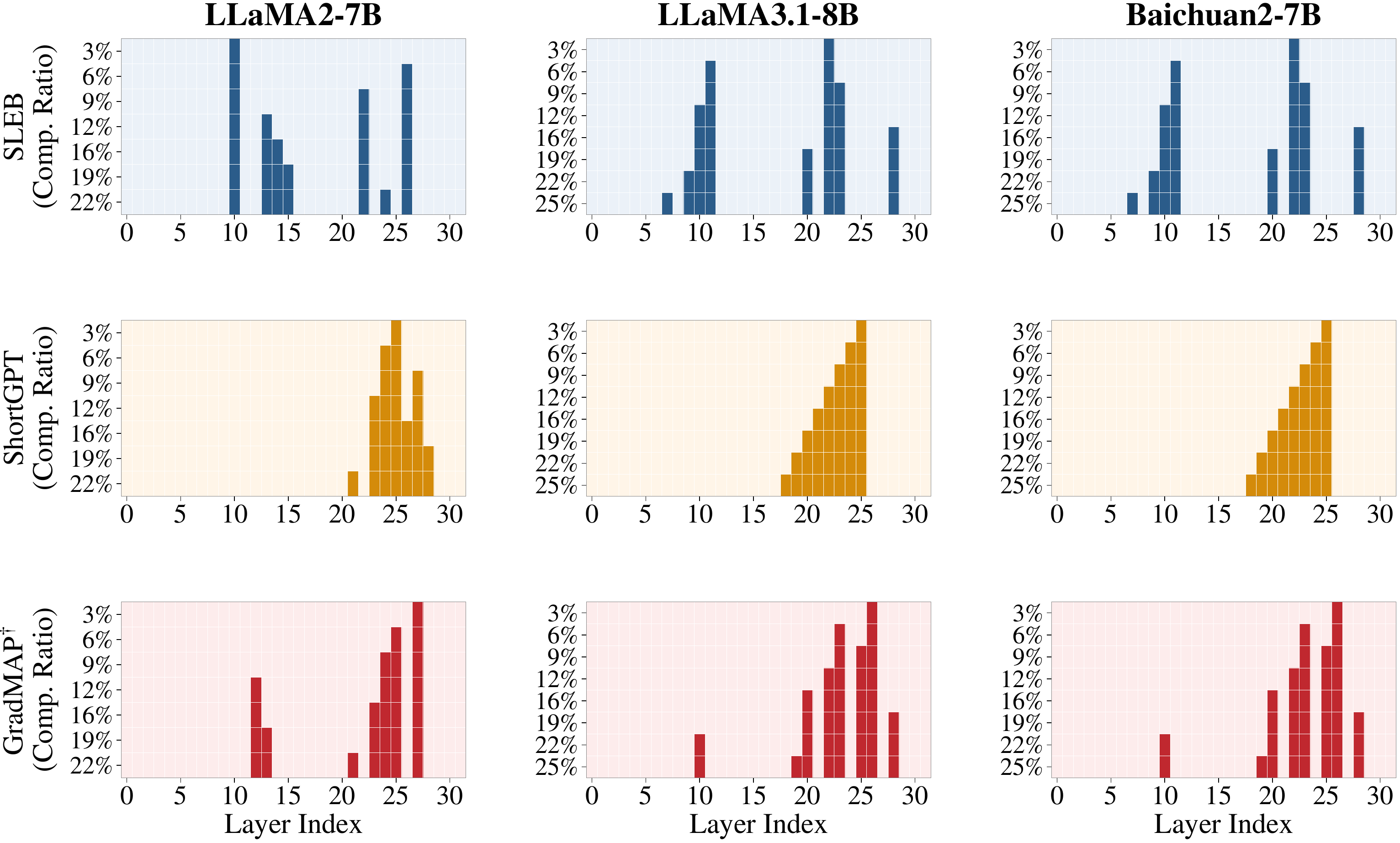}
    \caption{Comparison of layer selection strategies for GradMAP and other baselines across different compression ratios.}
    \label{fig:layer_selection}
\end{figure}

\subsection{Experimental Results}
\textbf{Zero-Shot Performance.} 
We conducted model compression experiments across various architectures, with the main results showing the performance on LLaMA2-7B, LLaMA2-13B and Vicuna-7B.
The results are presented in Table~\ref{tab:zeroshot}. 
GradMAP achieves a $4\times$ acceleration in pruning time on average compared to existing methods while attaining best performance across all evaluated benchmarks. 
Notably, while MKA prunes in similar time, its models show a severe perplexity collapse at the same compression ratios, whereas GradMAP maintains strong performance.
For instance, with the LLaMA2-7B model, GradMAP maintains the PPL of 20.56, whereas MKA suffers catastrophic model collapse, with PPL exceeding 877.70.
In addition, we evaluate the robustness of the model by measuring the perplexity at different compress levels, as shown in Figure~\ref{fig:diffent-ratio}. GradMAP consistently outperforms all baselines across compress levels, with particularly significant gains in high compress ratio.
Figure~\ref{fig:diffent-ratio} shows perplexity across compression levels, where GradMAP consistently outperforms all baselines, with larger gains at higher compression ratios.
To evaluate generalization, we test our framework on seven commonsense reasoning benchmarks.
As shown in Table~\ref{tab:zeroshot}, GradMAP consistently outperforms all baselines across models at 25\% compression, demonstrating strong effectiveness and generality.

\begin{table*}[h]
\centering
\setlength\tabcolsep{10pt}
\footnotesize 
\resizebox{\textwidth}{!}{
\begin{tabular}{c|cccccc}
\toprule
\textbf{Method} & \textbf{MMLU}$\uparrow$ & \textbf{CMMLU}$\uparrow$ & \textbf{GSM8k}$\uparrow$ & \textbf{XSum}$\uparrow$ & \textbf{StrategyQA}$\uparrow$ & \textbf{Avg.}$\uparrow$ \\
\midrule
SLEB & 25.42 & 22.36 & 0.89 & 0.15 & 18.47 & 13.46 \\
ShortGPT & 28.59 & 40.62 & 7.89 & 1.67 & 20.13 & 19.78 \\
\textbf{\method\textsuperscript{\dag}} & \textbf{30.77} & 48.89 & 11.45 & 1.67 & 40.74 & 26.70 \\
\textbf{\method\textsuperscript{\ddag}} & 30.17 & \textbf{51.84} & \textbf{11.51} & \textbf{2.12} & \textbf{40.79} & \textbf{27.20} \\
\bottomrule
\end{tabular}
}
\centering
\caption{Performance comparison on additional datasets and generative tasks using Baichuan2-7B.}
\label{tab:baichuan_additional}
\end{table*}

\begin{figure}
    \centering
    \includegraphics[width=\linewidth]{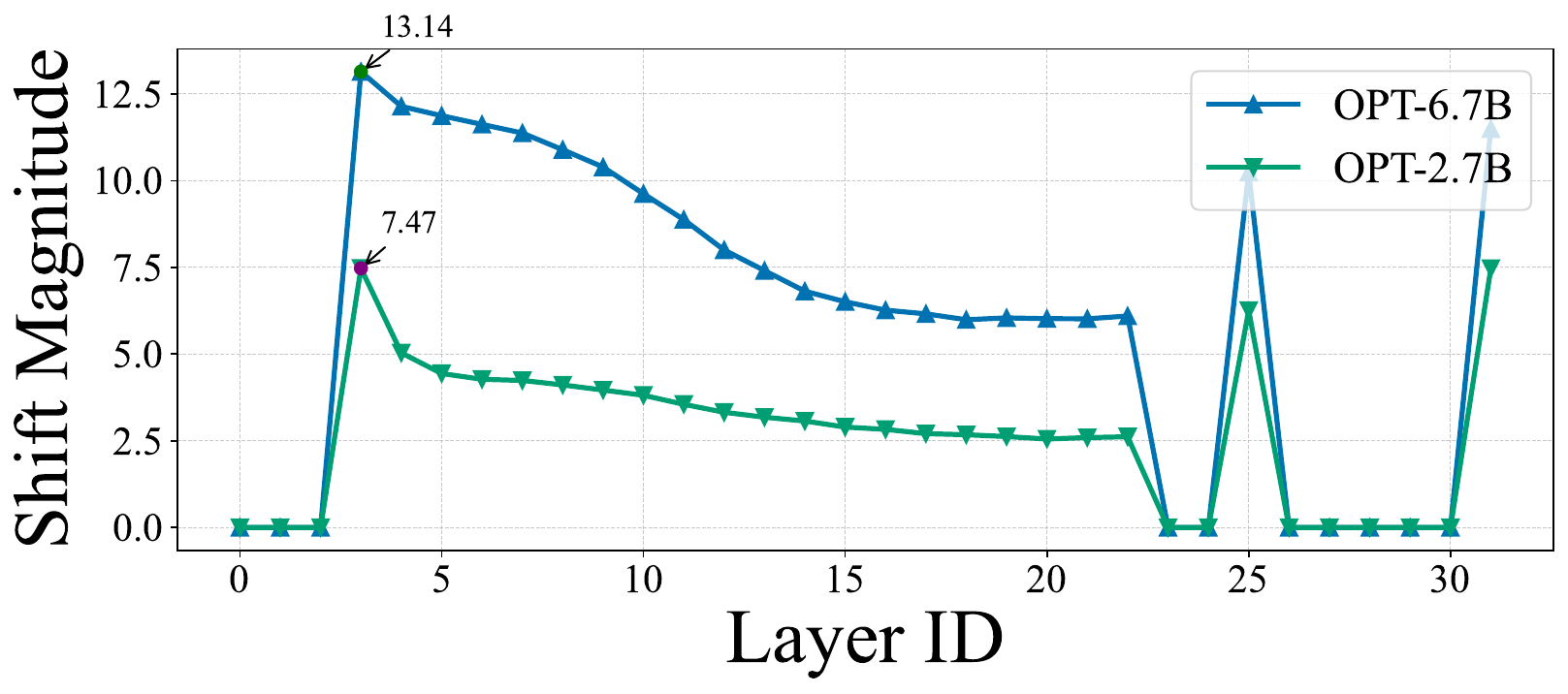}
    \caption{Distribution of first-order moment offsets before and after pruning for OPT-6.7B and OPT-2.7B.} \label{fig:shift_opt}
\end{figure}

\textbf{Generalization to More Models and Benchmarks.}
Table~\ref{tab:new_model} presents results on multiple architectures, including LLaMA3.1-8B, Baichuan2-7B, and OPT-6.7B, evaluated across a diverse set of benchmarks. 
Following the evaluation protocol of LLM-Streamline~\cite{LLM-Streamline}, we adopt OpenCompass~\cite{contributors2023opencompass} to conduct a comprehensive assessment on 12 tasks spanning natural language understanding and question-answering, including C3, CMNLI, CHID, WSC, HellaSwag, PIQA, RACE, MMLU, CMMLU, and CommonsenseQA. 

As shown in Table~\ref{tab:new_model}, GradMAP consistently outperforms competing methods across the majority of benchmarks and achieves the highest average accuracy on all evaluated models, demonstrating strong generalization across both model architectures and task distributions.


\textbf{Statistics of the Compressed Model.} Table~\ref{tab:throughput_latency} summarizes the inference throughput and latency of models compressed by different pruning strategies. Unstructured pruning methods show almost no inference speedup despite high compress ratio, while structured pruning provides moderate improvements as the compression ratio increases. 
In contrast, our layer pruning achieves the best efficiency, with 1.27$\times$ higher throughput and 1.26$\times$ lower latency at 20\% compression.

\begin{figure}
    \centering
    \includegraphics[width=\linewidth]{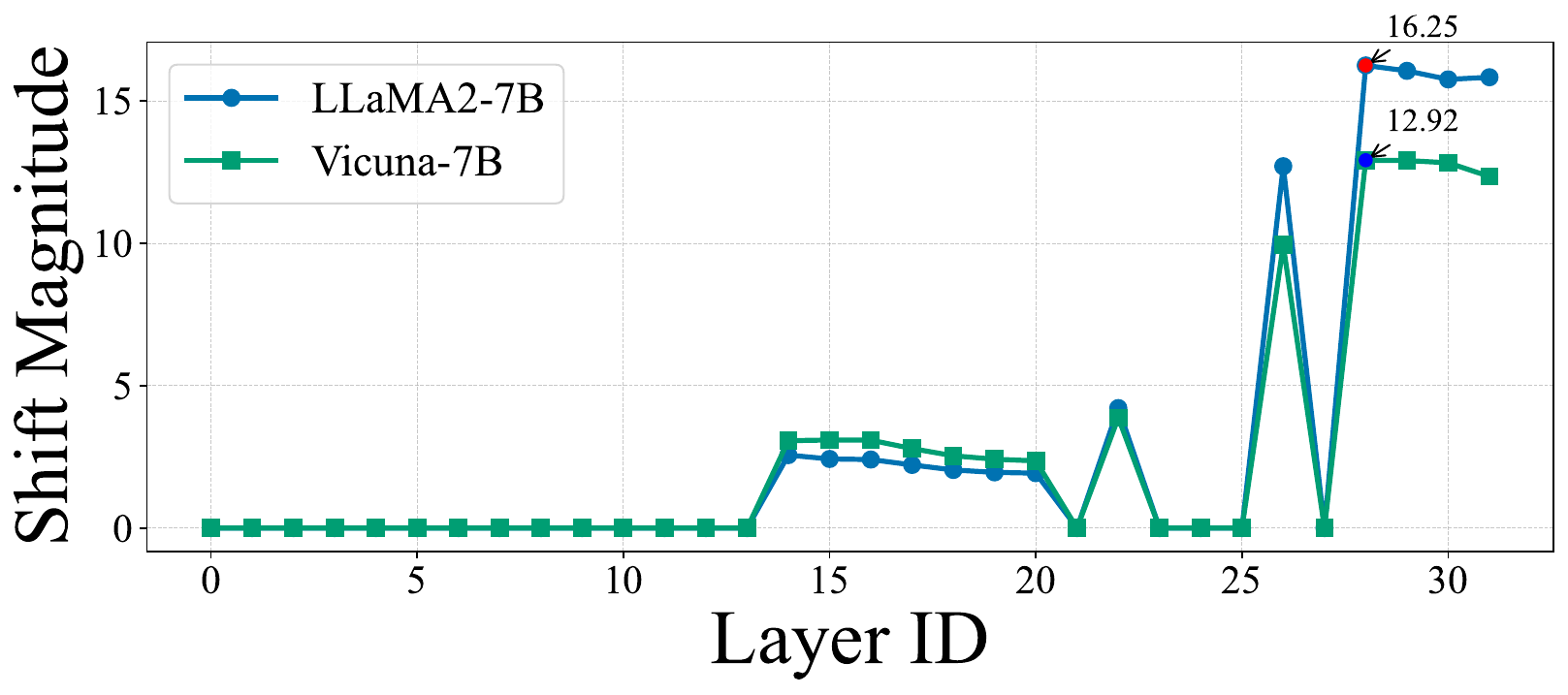}
    \caption{Distribution of first-order moment offsets before and after pruning for LLaMA2-7B and Vicuna-7B.} \label{fig:shift_llama2_vicuna}
\end{figure}

\textbf{Memory and Computational Cost.}
We further analyze the computational overhead of GradMAP. 
As shown in Figure~\ref{fig:memory_time}, both stages are lightweight in terms of memory and runtime. 
Stage~1 completes within a few minutes with moderate memory usage (e.g., about 26\,GB for LLaMA2-13B and less than 17\,GB for 7B-scale models), while Stage~2 is more efficient, finishing within one minute on a single NVIDIA A40 GPU with memory usage below 20\,GB for 7B models and within 30\,GB for larger models. 
These results indicate that GradMAP is efficient and practical for real-world deployment.

\textbf{Layer Selection Analysis.}
To better understand the pruning behavior of different methods, we visualize the layer selection patterns across models in Figure~\ref{fig:layer_selection}. ShortGPT tends to remove a contiguous block of layers concentrated in the latter half of the network, which may over-prune a specific functional region and harm the model's representational capacity. SLEB selects layers in a more scattered manner but still exhibits clustering around certain regions. In contrast, GradMAP selects layers that are more broadly distributed across the entire network depth, covering both middle and later layers. This dispersed selection pattern reflects the advantage of our gradient-informed importance metric, which evaluates each layer's contribution globally rather than relying on local heuristics. By avoiding excessive removal from any single region, GradMAP better preserves the overall information flow through the network, which contributes to its superior downstream performance.

\begin{figure*}
\centering
\subfloat[Accuracy on Baichuan2-7B]{
    \includegraphics[width=0.31\textwidth]{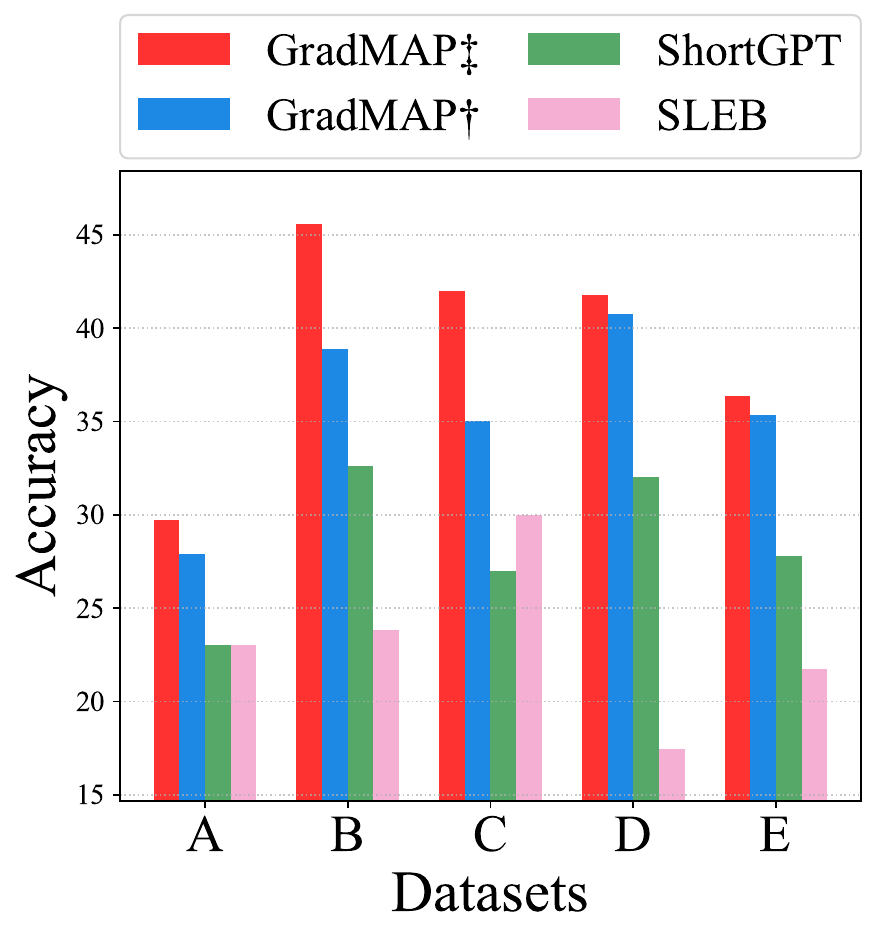}
}
\subfloat[Accuracy on Qwen2.5-7B]{
    \includegraphics[width=0.31\textwidth]{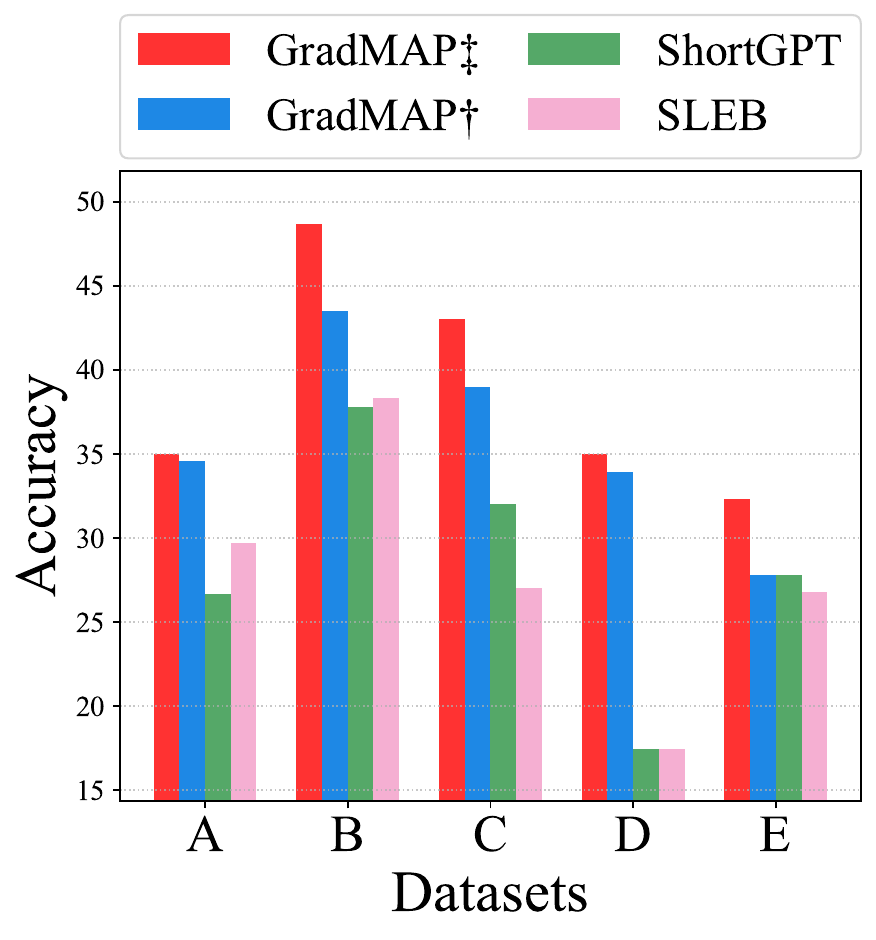}
}
\subfloat[Accuracy on LLaMA3.1-8B]{
    \includegraphics[width=0.31\textwidth]{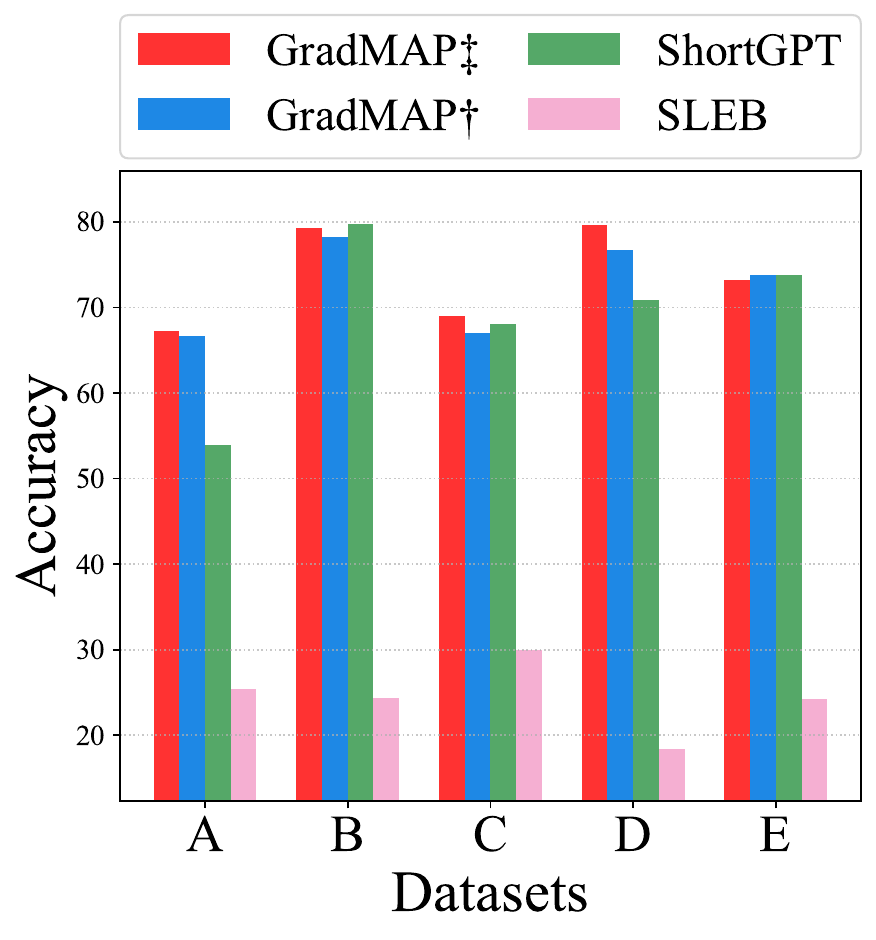}
}
\caption{
Accuracy comparison on different MMLU dataset subjects during pruning. 
Subjects: A: High School European History, B: High School Government and Politics, 
C: Medical Genetics, D: Management, E: High School Geography.
}
\label{fig:acc-mmlu}
\end{figure*}

\begin{figure}
    \centering
    \includegraphics[width=\linewidth]{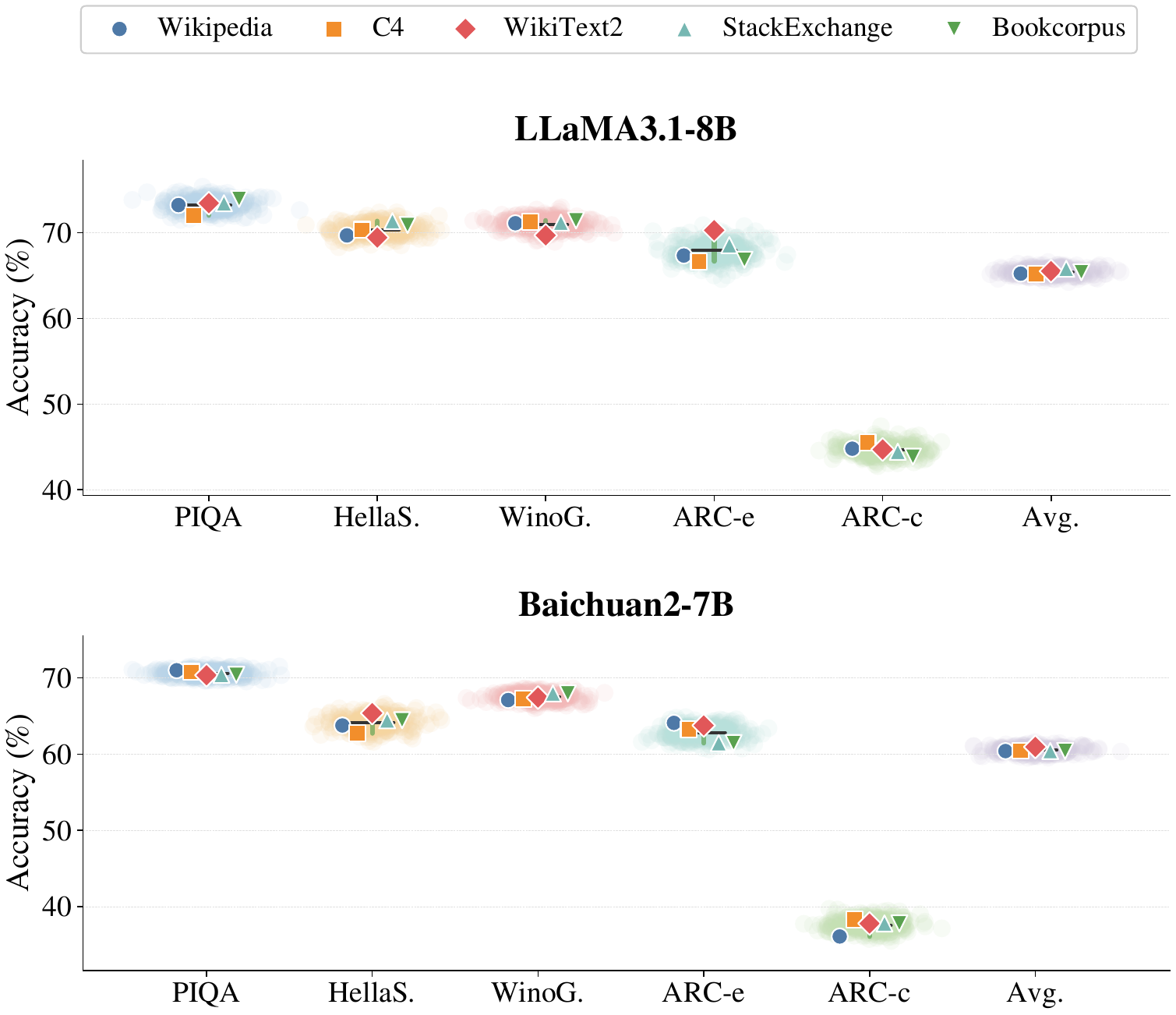}
    \vspace{-1mm}
    \caption{Performance comparison of GradMAP using different calibration datasets across multiple benchmarks.}
    \label{fig:calibration_deviation}
    \vspace{-1mm}
\end{figure}

\textbf{Projection Compensation Matrix Seamlessly Enhances Other Layer Pruning Method.}
Our projection compensation matrix can be easily integrated into existing layer pruning methods to restore performance.
We apply it to pruned models from several baselines and evaluate the effectiveness on the common-sense reasoning tasks. Results are shown in Table~\ref{tab:acc_withstage2}. 
After applying the projection compensation matrix, the methods show noticeable improvements.
These results demonstrate the effectiveness and generalizability of our compensation approach.
Notably, even with these enhancements, GradMAP still achieves superior performance, further validating the strength of our proposed metric.

\textbf{Results on Generative and Reasoning Tasks. }
To further demonstrate the robustness and generalization of our method, we additionally evaluate GradMAP on several reasoning and generative benchmarks, including GSM8k, XSum, and StrategyQA, using Baichuan2-7B as the backbone model. These tasks cover diverse evaluation settings, ranging from mathematical reasoning to abstractive summarization and strategic question answering. Table~\ref{tab:baichuan_additional} summarizes the performance comparison with representative baselines. Overall, GradMAP consistently achieves strong performance across all tasks.

\begin{table}[t]
\centering
\footnotesize
\setlength\tabcolsep{7pt}
\begin{tabular}{c|c|cc}
\toprule
\textbf{Component} & \textbf{Variant} & \textbf{LLaMA2-7B} & \textbf{Vicuna-7B} \\
\midrule

\multirow{2}{*}{Layer Selection}
& One-Shot & 50.07 & 50.92 \\
& \textbf{\method\textsuperscript{\dag}} & \textbf{56.08} & \textbf{55.88} \\

\midrule

\multirow{4}{*}{Compensation}
& Local & 56.66 & 55.57 \\
& Top-5 & 57.95 & 56.31 \\
& Top-2 & 58.21 & 57.70 \\
& \textbf{\method\textsuperscript{\ddag}} & \textbf{58.59} & \textbf{58.15} \\

\bottomrule
\end{tabular}
\caption{
Ablation study of layer selection and compensation strategies.
}
\label{tab:ablation_stage1_stage2}
\end{table}

\textbf{Layer-Wise Shift and Compensation Analysis. } 
We conducted a visual analysis of the first-order moment for retained layers across different models before and after pruning. The results are illustrated in Figure~\ref{fig:shift_opt} and Figure~\ref{fig:shift_llama2_vicuna}.
We observed an intriguing phenomenon regarding the locations of maximum shift magnitudes across different model architectures. As illustrated in Figure~\ref{fig:shift_llama2_vicuna}, for LLaMA2-7B and Vicuna-7B, the shift magnitudes gradually increase as layers are pruned; however, a decreasing trend emerges in the final layers, with the maximum offsets appearing notably at the (N-3)-th layer. In contrast, Figure~\ref{fig:shift_opt} shows that for the OPT model family, the shift magnitudes exhibit a consistent pattern of first increasing and then decreasing throughout the middle layers, with their peak occurring distinctly at the third layer. These findings highlight model differences in parameter sensitivities and suggest considerations for optimizing pruning strategies across various LLMs.

\textbf{Results on MMLU Subjects.}
We conduct zero-shot evaluations on representative subjects from the MMLU benchmark using Baichuan2-7B, Qwen2.5-7B, and LLaMA3.1-8B. The selected subjects span diverse domains, including humanities, social sciences, and professional knowledge, providing a comprehensive evaluation of model capabilities under pruning. The results are shown in Figure~\ref{fig:acc-mmlu}. 
Overall, GradMAP consistently outperforms the baselines across different subjects and model architectures. In particular, it maintains more stable accuracy as the compression ratio increases, effectively mitigating performance degradation caused by pruning. The improvements are especially evident on knowledge-intensive and reasoning-related tasks, indicating that GradMAP better preserves essential knowledge and reasoning ability during compression.


\begin{table*}[t]
\centering
\footnotesize
\setlength{\tabcolsep}{5pt}
\resizebox{\textwidth}{!}{\begin{tabular}{c|c|ccccccc|c}
\toprule
\textbf{Types} 
& \textbf{PPL}$\downarrow$ &
\textbf{BoolQ}$\uparrow$ & \textbf{PIQA}$\uparrow$ & \textbf{HellaS.}$\uparrow$ & \textbf{WinoG.}$\uparrow$ & \textbf{ARC-e}$\uparrow$ & \textbf{ARC-c}$\uparrow$ & \textbf{OBQA}$\uparrow$ & \textbf{Avg.}$\uparrow$ \\
\midrule
\rowcolor{lightgray}
\multicolumn{10}{c}{\textit{MHA Weights}} \\
$W_Q$    
& 28.49 & 50.98 & 70.73 & 61.17 & 61.80 & 63.01 & 37.80 & 38.20 & 54.81 \\
$W_K$    
& 28.49 & 50.98 & 70.73 & 61.17 & 61.80 & 63.01 & 37.80 & 38.20 & 54.81 \\
$W_V$    
& 28.54 & 51.65 & 70.51 & 61.10 & 61.80 & \textbf{63.09} & 38.31 & 38.20 & 54.95 \\
$W_O$    
& 27.94 & 56.36 & 70.67 & 61.00 & 61.72 & \textbf{63.09} & 37.88 & 37.40 & 55.45 \\
\midrule
\rowcolor{lightgray}
\multicolumn{10}{c}{\textit{FFN Weights}} \\
$W_{\text{up}}$ 
& 29.05 & 49.63 & 70.18 & 60.77 & 61.96 & 62.58 & 37.46 & 38.00 & 54.37 \\
$W_{\text{gate}}$ 
& 29.68 & 54.74 & 70.13 & 60.91 & \textbf{62.19} & 62.37 & 37.88 & 37.20 & 55.06 \\
$W_{\text{down}}$ 
& \textbf{27.56} & \textbf{72.79} & \textbf{70.89} & \textbf{61.32} & 61.48 & 63.01 & \textbf{38.65} & \textbf{38.80} & \textbf{58.06} \\
\bottomrule
\end{tabular}}
\caption{Ablation of compensation on different weight matrices for Vicuna-7B at 25\% compression ratio.}
\label{tab:ablation_weight_matrix}
\end{table*}


\begin{table}[!ht]
  \centering
    \footnotesize
    \setlength\tabcolsep{15pt}
  \begin{tabular}{c|cc|c}
    \toprule
    \textbf{Model} & $\boldsymbol{\mathcal{L}_{\mathrm{MSE}}}$ & $\boldsymbol{\mathcal{L}_{\mathrm{reg}}}$ & \textbf{Avg.$\uparrow$} \\
    \midrule
    \multirow{3}{*}{{\textbf{LLaMA2-7B}}} 
    & \cmark & \xmark  & 56.54 \\
    & \xmark & \cmark  & 56.66 \\
    & \cmark & \cmark  & \textbf{58.59} \\
    \midrule
    \multirow{3}{*}{{\textbf{Vicuna-7B}}}
    & \cmark & \xmark  & 56.98 \\
    & \xmark & \cmark  & 55.32 \\
    & \cmark & \cmark  & \textbf{58.15} \\
    \bottomrule
  \end{tabular}
      \caption{Ablation study of loss functions.  }
    \label{tab:loss-ablation}
\end{table}

\subsection{Ablation Study}
\label{sec:ablation-study}
\textbf{Iterative Pruning and OneShot Pruning Analysis.} 
We compared iterative and OneShot searches for unimportant layers in Stage 1 of GradMAP. As Table~\ref{tab:ablation_stage1_stage2} shows, OneShot Pruning accelerates the process but reduces performance: for LLaMA2-7B, iterative pruning achieved 56.08\% accuracy versus 50.07\% with OneShot, a drop of 6.01\%. We analyze that although our proposed metric can identify unimportant layers, the dependencies between layers in the model's outputs require an iterative approach to update the importance scores. Since our metric is computationally efficient, the time spent remains within an acceptable range even with the iterative approach. 
Notably, unlike SLEB, GradMAP makes each pruning decision using a single gradient computation rather than exhaustively masking and evaluating all candidates.

\textbf{The Different Compensation Strategies. }
We examined two strategies: local compensation, which adjusts neighboring layer weights (e.g., updating layer 2 when layer 3 is pruned), and GradMAP\textsuperscript{\ddag}, which identifies layers with the largest output shifts through a one-time analysis and selectively adjusts them. As Table~\ref{tab:ablation_stage1_stage2} shows, GradMAP\textsuperscript{\ddag} substantially improves performance. Ablation on Top-$Z$ shift layers reveals diminishing or negative returns as $Z$ increases, likely due to overfitting limited calibration data and eroding pre-trained knowledge, while larger $Z$ adds unnecessary computation. We therefore adopt a minimal strategy, adjusting only the most drifted layer.

\begin{table}[!ht]
  \centering
  \setlength\tabcolsep{10pt}
  \footnotesize
  \begin{tabular}{c|c|cc}
    \toprule
    \textbf{Model} & \textbf{Type} & \textbf{Time}$\downarrow$ & \textbf{Accuracy}$\uparrow$ \\
    \midrule
    \multirow{2}{*}{\textbf{LLaMA2-7B}} 
    & Right  & 1475.1 & \textbf{58.70} \\
    & Left & \textbf{54.0}   & 58.59 \\
    \midrule
    \multirow{2}{*}{\textbf{LLaMA2-13B}} 
    & Right  & 2758.3 & 59.48 \\
    & Left & \textbf{93.1}   & \textbf{59.49} \\
    \midrule
    \multirow{2}{*}{\textbf{Vicuna-7B}} 
    & Right  & 1487.2 & 58.05 \\
    & Left & \textbf{80.6}   & \textbf{58.15} \\
    \bottomrule
  \end{tabular}
      \caption{Comparison of different calibration types on zero-shot performance.}
  \label{tab:ablation_calibrate_type}
\end{table}

\textbf{The Impact of Loss Function. }
We perform an ablation study on the loss components of our projection compensation matrix learning by analyzing the individual contributions of the mean squared error (MSE) loss and the regularization term in our objective function.
The quantitative results are presented in Table~\ref{tab:loss-ablation}.
Our findings indicate that using MSE loss alone leads to severe overfitting on the calibration dataset, as the model fails to generalize beyond the limited 128 calibration samples. 
On the other hand, relying solely on regularization loss fails to provide meaningful performance improvements, as the projection compensation matrix lacks sufficient adaptation to the pruned model.
At the 25.00\% compression ratio on Vicuna-7B, using only MSE loss yields 56.98\% accuracy, and only regularization loss gives 55.32\%. In contrast, our combined loss achieves 58.15\%, outperforming MSE by 2.0\% and regularization by 5.1\%. This highlights the effectiveness of our loss design in stabilizing pruned model performance. Therefore, we adopt a balanced combination of MSE and regularization to ensure effective weight adaptation and model robustness after pruning.

\textbf{The Impact of Calibration Type.}
We investigate different strategies for applying the projection compensation matrix. Specifically, we examine the effects of either left-multiplying or right-multiplying the down-projection matrix within the FFN. The results of these two approaches are summarized in Table~\ref{tab:ablation_calibrate_type}.
For the weight matrix, we denote its dimension as $\mathbf{W}^{(i^*)} \in \mathbb{R}^{d \times k}$, with the condition that $k \gg d$.
Consequently, training a left-multiplying projection compensation matrix requires learning parameters in a matrix of dimension $\mathbf{W_{left}}' \in \mathbb{R}^{d \times d}$, whereas a right-multiplying projection compensation matrix necessitates training parameters in a matrix of dimension $\mathbf{W_{right}}' \in \mathbb{R}^{k \times k}$. 
Therefore, employing a left-multiplying projection compensation matrix significantly reduces computational overhead, leading to faster and more efficient training. 
Experimental results corroborate this efficiency, demonstrating that utilizing a left-multiplying projection compensation matrix maintains model performance while achieving a \textbf{20-fold} increase in calibrating speed.

\textbf{Robustness to Calibration Data.}
We investigate the sensitivity of GradMAP to the choice of calibration dataset. As shown in Figure~\ref{fig:calibration_deviation}, we evaluate five representative calibration sources, including Wikipedia, C4, WikiText2, StackExchange, and BookCorpus, across multiple common-sense reasoning benchmarks on both LLaMA3.1-8B and Baichuan2-7B. These datasets differ significantly in domain, style, and linguistic characteristics, providing a comprehensive test of robustness.
The results show that GradMAP maintains consistently stable performance across all calibration datasets, with only marginal variations observed on different benchmarks and model architectures. 
In contrast to methods that rely on activation statistics and are therefore sensitive to data distribution, GradMAP leverages a gradient-based importance metric that reflects the intrinsic structural contribution of each layer. This property enables it to remain effective even when the calibration data deviates from the target task distribution.
Overall, these findings demonstrate that GradMAP is largely insensitive to the choice of calibration data, reducing the dependency on carefully curated datasets and making the method more practical and reliable in real-world deployment scenarios.

\textbf{The Compensation on different weight matrices.} 
We conduct an ablation study by applying the compensation module to different weight matrices in both MHA and FFN on Vicuna-7B at a 25\% compression ratio (Table~\ref{tab:ablation_weight_matrix}).
For MHA, compensation is less effective, as the projection matrices are relatively distant from the final output and are entangled with non-linear operations such as softmax. This indirect influence weakens the ability of a linear compensation module to accurately correct the output distribution, leading to limited performance gains.
In contrast, the FFN structure exhibits clearer distinctions. Among its components, only $W_{\text{down}}$ serves as a direct linear mapping to the output space, while $W_{\text{up}}$ and $W_{\text{gate}}$ are embedded within non-linear activation functions. As a result, applying linear compensation to $W_{\text{up}}$ and $W_{\text{gate}}$ tends to be unstable and less effective, since their influence on the final output is highly non-linear and input-dependent.
These observations indicate that the effectiveness of compensation is closely related to the linearity and positional proximity of the target weight matrix to the model output. Consequently, compensating $W_{\text{down}}$ achieves the best performance, yielding the highest average accuracy across all benchmarks. This also validates our design choice of focusing the compensation module on the most structurally suitable component within the FFN.


\section{Conclusion}
\label{Conclusions}
In this paper, we propose GradMAP, a novel layer pruning and weight compensation algorithm for LLMs. 
Our method addresses two critical challenges in layer pruning, namely efficient importance estimation and effective performance recovery. 
Specifically, we introduce a gradient-based importance metric that leverages global gradient magnitudes to identify unimportant layers with only a single backward pass, significantly improving pruning efficiency. 
To further mitigate performance degradation, we propose a projection compensation matrix that aligns the pruned model’s outputs with those of the original model by correcting the first-order moment shift in activations. 
This two-stage design enables GradMAP to achieve a strong balance between efficiency and accuracy.
Extensive experiments across multiple LLMs and benchmarks demonstrate that GradMAP consistently outperforms existing layer pruning methods in both effectiveness and efficiency, achieving comparable or better performance while accelerating the pruning process by an average factor of $4\times$. 
These results highlight the potential of gradient-based metrics and lightweight compensation mechanisms for LLM compression.

\section{Acknowledgement}
This work was supported  by the National Natural Science Foundation of China (62476274, U22B2048, 62394330).

\bibliographystyle{cas-model2-names}
\bibliography{refer}

\end{document}